\documentclass{article}

    \PassOptionsToPackage{numbers, compress}{natbib}

\usepackage[final]{neurips_2022}

\usepackage[utf8]{inputenc} 
\usepackage[T1]{fontenc}    
\usepackage{hyperref}       
\usepackage{url}            
\usepackage{booktabs}       
\usepackage{amsfonts}       
\usepackage{nicefrac}       
\usepackage{microtype}      
\usepackage{xcolor}         

\usepackage{graphicx}
\usepackage{amsmath}
\usepackage{multirow}
\usepackage[ruled,vlined]{algorithm2e}
\usepackage{booktabs}
\usepackage{tablefootnote}
\usepackage{amsthm}

\SetAlFnt{\footnotesize}
\SetAlCapFnt{\small}
\SetAlCapNameFnt{\small}
\SetAlCapHSkip{0em}
\SetAlCapSkip{0ex}
\newtheorem{proposition}{Proposition}
\newtheorem{remark}{Remark}

\renewcommand{\vec}{\mathbf}
\newcommand{\mrm}{\mathrm}

\newcommand\sbullet[1][.5]{\mathbin{\vcenter{\hbox{\scalebox{#1}{$\bullet$}}}}}
\newcommand\sig{$\sbullet[.5]$ }
\newcommand\ssig{$\sbullet[.75]$ }
\newcommand\sssig{$\sbullet[1.]$ }

\title{SPD domain-specific batch normalization to crack interpretable unsupervised domain adaptation in EEG}

\author{%
  Reinmar J.~Kobler$^{1,2}$, Jun-ichiro Hirayama$^{1}$, Qibin Zhao$^{1}$, Motoaki Kawanabe$^{1,2}$ \\
  $^{1}$RIKEN Center for Advanced Intelligence Project (RIKEN AIP), Tokyo, Japan \\
  $^{2}$Advanced Telecommunications Research Institute International (ATR), Kyoto, Japan \\
  \texttt{kobler.reinmar@gmail.com, qibin.zhao@riken.jp}\\\texttt{jun-ichiro.hirayama@a.riken.jp},~\texttt{kawanabe@atr.jp} \\
}

\begin{document}

\maketitle

\begin{abstract}

Electroencephalography (EEG) provides access to neuronal dynamics non-invasively with millisecond resolution, rendering it a viable method in neuroscience and healthcare.
However, its utility is limited as current EEG technology does not generalize well across domains (i.e., sessions and subjects) without expensive supervised re-calibration.
Contemporary methods cast this transfer learning (TL) problem as a multi-source/-target unsupervised domain adaptation (UDA) problem and address it with deep learning or shallow, Riemannian geometry aware alignment methods.
Both directions have, so far, failed to consistently close the performance gap to state-of-the-art domain-specific methods based on tangent space mapping (TSM) on the symmetric, positive definite (SPD) manifold.
Here, we propose a machine learning framework that enables, for the first time, learning domain-invariant TSM models in an end-to-end fashion.
To achieve this, we propose a new building block for geometric deep learning, which we denote  SPD domain-specific momentum batch normalization (SPDDSMBN).
A SPDDSMBN layer can transform domain-specific SPD inputs into domain-invariant SPD outputs, and can be readily applied to multi-source/-target and online UDA scenarios.
In extensive experiments with 6 diverse EEG brain-computer interface (BCI) datasets, we obtain state-of-the-art performance in inter-session and -subject TL with a simple, intrinsically interpretable network architecture, which we denote TSMNet.
Code: \url{https://github.com/rkobler/TSMNet}

\end{abstract}

\section{Introduction}

Electroencephalography (EEG) measures multi-channel electric brain activity from the human scalp with millisecond precision \cite{Schomer2011}.
Transient modulations in the rhythmic brain activity can reveal cognitive processes \cite{pfurtscheller_event-related_1999}, affective states \cite{faller_regulation_2019} and a person's health status \cite{zhang_identification_2021}.
Unfortunately, these modulations exhibit low signal-to-noise ratio (SNR), domain shifts (i.e., changes in the data distribution) and have low specificity, rendering statistical learning a challenging task - particularly in the context of brain-computer interfaces (BCI) \cite{Wolpaw2002} where the goal is to predict a target from a short segment of multi-channel EEG data in real-time.

Under domain shifts, domain adaptation (DA), defined as learning a model from a source domain that performs well on a related target domain, offers principled statistical learning approaches with theoretical guarantees \cite{ben-david_theory_2010,hoffmann_algorithms_2018}.
DA in the BCI field mainly distinguishes inter-session and -subject transfer learning (TL) \cite{wu_transfer_2020}.
In inter-session TL, domain shifts, are expected across sessions mainly due to mental drifts (low specificity) as well as differences in the relative positioning of the electrodes and their impedances.
Inter-subject TL is more difficult, as domain shifts are additionally driven by structural and functional differences in brain networks as well as variations in the performed task \cite{neuper_imagery_2005}.

These domain shifts are traditionally circumvented by recording labeled calibration data and fitting domain-specific models \cite{lotte_review_2018,Statthaler2017}. 
As recording calibration data is costly, models that are robust to scarce data with low SNR perform well in practice.
Currently, tangent space mapping (TSM) models \cite{Barachant2012,sabbagh_manifold-regression_2019} operating with symmetric, positive definite (SPD) covariance descriptors of preprocessed data are considered state-of-the-art (SoA) \cite{lotte_review_2018,jayaram_moabb_2018,roy_retrospective_2022}.
They are well suited for EEG data as they exhibit invariances to linear mixing of latent sources \cite{congedo_riemannian_2017}, and are consistent \cite{sabbagh_manifold-regression_2019} and intrinsically interpretable \cite{kobler_interpretation_2021} estimators for generative models that encode label information with a log-linear relationship in source power modulations.

Competitive, supervised calibration-free methods are one of the long-lasting grand challenges in EEG neurotechnology research \cite{Wolpaw2002,lotte_review_2018,fairclough_grand_2020,wei_2021_2022,roy_retrospective_2022}.
Among the applied transfer learning techniques, including multi-task learning \cite{jayaram_transfer_2016} and domain-invariant learning \cite{Fazli2009,Samek2014,kwon_subject-independent_2020}, unsupervised domain adaptation (UDA) \cite{zhao_multi-source_2020} is considered as key to overcome this challenge \cite{lotte_review_2018,wei_2021_2022}.
Contemporary methods cast the problem as a multi source and target UDA problem and address it with deep learning \cite{ozdenizci_learning_2020,xu_improving_2021,xu_crossdataset_2020,mane_fbcnet_2021} or shallow, Riemannian geometry aware alignment methods \cite{zanini_transfer_2018,rodrigues_riemannian_2019,yair_parallel_2019,yong_momentum_2020}.
Successful methods must cope with notoriously small and heterogeneous datasets (i.e., dozens of domains with a few dozens observations per domain and class).
In a recent, relatively large scale inter-subject and -dataset TL competition with few labeled examples per target domain \cite{wei_2021_2022}, deep learning approaches that aligned the first and second order statistics either in input \cite{he_transfer_2020,xu_crossdataset_2020} or latent space \cite{bakas_team_2022} obtained the highest scores.
Whereas, in a pure UDA competition \cite{roy_retrospective_2022} with a smaller dataset, Riemannian geometry aware approaches dominated.
With the increasing popularity of geometric deep learning \cite{bronstein_geometric_2017}, \citet{ju_deep_2022} proposed an architecture based on SPD neural networks \cite{huang_riemannian_2017} to align SPD features in latent space and attained SoA scores.
Despite the tremendous advances in recent years, the field still lacks methods that can consistently close the performance gap to state-of-the-art domain-specific methods.

\begin{figure}

  \centering
  \includegraphics[width=\textwidth]{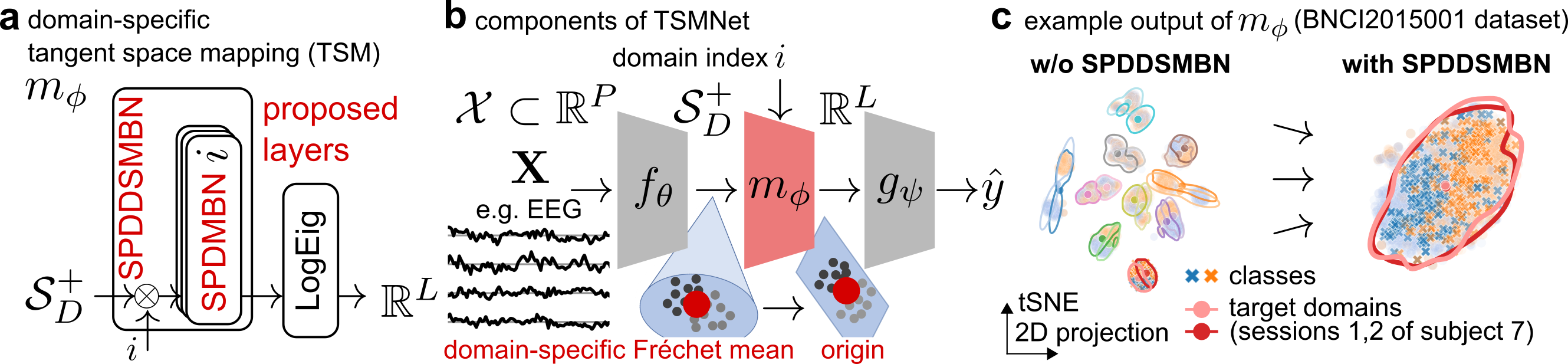}
  \caption{Visualization of the proposed framework around (\textbf{a}) SPD domain-specific momentum batch normalization (SPDDSMBN) that (\textbf{b}) learns parameters $\Theta=\{\theta,\phi, \psi \}$ of typical tangent space mapping (TSM) models \emph{end-to-end} to crack multi-source/-target unsupervised domain adaptation on $\mathcal{S}^+_D$ for EEG data (\textbf{c}, illustrative example).
  For EEG data, we propose a simple, intrinsically interpretable parametrization of $f$ and $g$, denoted TSMNet, and obtain SoA performance. 
  }\label{fig:overview}
\end{figure}

To close this gap, we propose a machine learning framework around domain-specific batch normalization on the SPD manifold (Figure\,\ref{fig:overview}).
The proposed framework is used to implement domain-specific TSM (Figure\,\ref{fig:overview}a), which requires tracking the domains' Fréchet means in latent space as they are changing during training a typical TSM model in an end-to-end fashion (Figure\,\ref{fig:overview}b).
After reviewing some preliminaries in section 2, we extend momentum batch normalization (MBN) \cite{yong_momentum_2020} to SPDMBN that controls the Fréchet mean and variance of SPD data in section 3.
In a theoretical analysis, we show under reasonable assumptions that SPDMBN can track and converge to the data's true Fréchet mean, enabling, for the first time, end-to-end learning of feature extractors, TSM and tangent space classifiers.
Building upon this insight, we combine SPDMBN with domain-specific batch normalization (DSBN) \cite{chang_domainspecific_2019} to form SPDDSMBN (Figure\,\ref{fig:overview}a).
A SPDDSMBN layer can transform domain-specific SPD inputs into domain-invariant SPD outputs (Figure\,\ref{fig:overview}c).
Like DSBN, SPDDSMBN easily extends to multi-source, multi-target and online UDA scenarios.
In section 4, we briefly review the generative model of EEG, before the proposed methods are combined in a simple, intrinsically interpretable network architecture, denoted TSMNet (Figure\,\ref{fig:overview}b).
We obtain state-of-the-art performance in inter-session and -subject UDA on small and large scale EEG BCI datasets, and show in an ablation study that the performance increase is primarily driven by performing DSBN on the SPD manifold.

\section{Preliminaries}
\label{sec:preliminaries}
\vspace{-.5em}
\paragraph{Multi-source multi-target unsupervised domain adaptation}
Let $\mathcal{X}$ denote the space of input features, $\mathcal{Y}$ a label space, and $\mathcal{I}_d \subset \mathbb{N}$ an index set that contains unique domain identifiers.
In the multi-source, multi-target unsupervised domain adaptation scenario considered here, we are given a set $\mathcal{T}^{source} = \{\mathcal{T}_i | i \in \mathcal{I}_d^{source} \subset \mathcal{I}_d\}$ with $|\mathcal{I}_d^{source}|=N$ domains.
Each domain $\mathcal{T}_i = \{(\vec{X}_{ij},y_{ij})\}_{j=1}^{M} \sim P_{XY}^i$ contains $M$ observations of feature ($\vec{X} \in \mathcal{X}$) and label ($y \in \mathcal{Y}$) tuples sampled from a joint distribution $P_{XY}^i$.\footnote{For ease of notation, although not required by our method, we assume that $M$ is the same for each domain.}
While the joint distributions can be different (but related) across domains, we assume that the class priors are the same (i.e., $P_{Y}^i = P_{Y}$).
The goal is to learn a predictive function $h : \mathcal{X}\times \mathcal{I}_d \rightarrow \mathcal{Y}$ that, once fitted to $\mathcal{T}^{source}$, can generalize to unseen target domains $\mathcal{T}^{target} = \{\mathcal{T}_l | l \in \mathcal{I}_d^{target} \subset \mathcal{I}_d, \mathcal{I}_d^{target} \cap \mathcal{I}_d^{source} = \emptyset\}$ merely based on \emph{unsupervised} adaptation of $h$ to each target domain $\mathcal{T}_l$ once its label $l$ and features $\{\vec{X}_{lj}\}_{j=1}^{M} \sim P_{X}^l$ are revealed.
\vspace{-.5em}
\paragraph{Riemannian geometry on $\mathcal{S}^+_D$}
We start with recalling notions of geometry on the space of real $D \times D$ symmetric positive definite (SPD) matrices $\mathcal{S}^+_D = \{ \vec{Z} \in \mathbb{R}^{D \times D} : \vec{Z}^T = \vec{Z}, \vec{Z} \succ 0 \}$.
The space $\mathcal{S}^+_D$ forms a cone shaped Riemannian manifold in $ \mathbb{R}^{D\times D} $ \cite{bhatia_positive_2009}.
A Riemannian manifold $\mathcal{M}$ is a smooth manifold equipped with an inner product on the tangent space $\mathcal{T}_\vec{Z}\mathcal{M}$ at each point $\vec{Z} \in \mathcal{M}$. 
Tangent spaces have Euclidean structure with easy to compute distances $\mathcal{T}_\vec{Z}\mathcal{M} \times \mathcal{T}_\vec{Z}\mathcal{M} \rightarrow \mathbb{R}^+$ which locally approximate Riemannian distances on $\mathcal{M}$ induced by an inner product \cite{absil_optimization_2009}.
Logarithmic $\mathrm{Log}_\vec{Z}:\mathcal{M} \rightarrow \mathcal{T}_\vec{Z}\mathcal{M}$ and exponential  $\mathrm{Exp}_\vec{Z}: \mathcal{T}_\vec{Z}\mathcal{M} \rightarrow \mathcal{M}$ mappings project points to and from tangent spaces.

Using the inner product $\langle\vec{S}_1,\vec{S}_2\rangle_\vec{Z} = \mathrm{Tr}(\vec{Z}^{-1} \vec{S}_1 \vec{Z}^{-1} \vec{S}_2 )$ for points $\vec{S}_1,\vec{S}_2$ in the tangent space $ \mathcal{T}_\vec{Z}\mathcal{S}^+_D$ (i.e., the space of real symmetric $D \times D$ matrices) results in a globally defined affine invariant Riemannian metric on $\mathcal{S}^+_D$ \cite{pennec_riemannian_2006,bhatia_positive_2009}, which can be computed in closed form:
 \begin{equation}
  \delta(\vec{Z}_1,\vec{Z}_2) = ||\mathrm{log}(\vec{Z}_1^{-\frac{1}{2}}\vec{Z}_2\vec{Z}_1^{-\frac{1}{2}})||_F
 \label{eq:airm}
 \end{equation}
 where $\vec{Z}_1$ and $\vec{Z}_2$ are two SPD matrices, $\mathrm{log}(\cdot)$ denotes the matrix logarithm\footnote{For SPD matrices, powers, logarithms and exponentials can be computed via eigen decomposition.}, $||\cdot||_F$ the Frobenius norm, and $\mathrm{Tr}(\cdot)$ in the inner product the trace operator.
 Due to affine invariance, we have $\delta_{}(\vec{A} \vec{Z}_1 \vec{A}^T, \vec{A}\vec{Z}_2 \vec{A}^T) = \delta_{}(\vec{Z}_1,\vec{Z}_2)$ for any invertible $D \times D$ transformation matrix $\vec{A}$.
 The exponential and logarithmic mapping are also globally defined in closed form as
 \begin{equation}
  \mathrm{Log}_\vec{Z}(\vec{Z}_1) = \vec{Z}^{\frac{1}{2}} \mathrm{log}(\vec{Z}^{-\frac{1}{2}}\vec{Z}_1\vec{Z}^{-\frac{1}{2}}) \vec{Z}^{\frac{1}{2}}
  \label{eq:log}
 \end{equation}
 \begin{equation}
  \mathrm{Exp}_\vec{Z}(\vec{S}_1) = \vec{Z}^{\frac{1}{2}} \mathrm{exp}(\vec{Z}^{-\frac{1}{2}}\vec{S}_1\vec{Z}^{-\frac{1}{2}}) \vec{Z}^{\frac{1}{2}}
  \label{eq:exp}
 \end{equation}

 For a set of SPD points $\mathcal{Z} = \{\vec{Z}_j \in \mathcal{S}^+_D\}_{j \le M}$, we will use the notion of Fréchet mean $\vec{G}_{\mathcal{Z}} \in \mathcal{S}^+_D $ and Fréchet variance $\nu^2_{\mathcal{Z}} \in \mathbb{R}^+$.
 The Fréchet mean is defined as the minimizer of the average squared distances
 \begin{equation}
  \vec{G}_{\mathcal{Z}} = \mathrm{arg \min_{\vec{G}\in \mathcal{S}^+_D}} \frac{1}{M}\sum_{j=1}^{M} \delta_{}^2(\vec{G},\vec{Z}_j)
 \label{eq:geommean}
 \end{equation}
 For $M=2$, there is a closed form solution expressed as
 \begin{equation}
  \vec{G}_{\mathcal{Z}}(\gamma) = \vec{Z}_1 \#_\gamma \vec{Z}_2 = \vec{Z}_1^{\frac{1}{2}} \left(\vec{Z}_1^{-\frac{1}{2}}\vec{Z}_2\vec{Z}_1^{-\frac{1}{2}}\right)^\gamma \vec{Z}_1^{\frac{1}{2}}
  \label{eq:mean2points}
 \end{equation}
 with weight $ \gamma = 0.5$. 
 Choosing $ \gamma \in [0,1]$ computes weighted means along the geodesic (i.e., the shortest curve) that connects both points.
 For $M > 2$, (\ref{eq:geommean}) can be solved using the Karcher flow algorithm \cite{karcher_riemannian_1977}, which iterates between projecting the data to the tangent space (\ref{eq:log}) at the current estimate, arithmetic averaging, and projecting the result back (\ref{eq:exp}) to obtain a new estimate.
 The Fréchet variance $\nu_{\mathcal{Z}}^2$ is defined as the attained value at the minimizer $\vec{G}_{\mathcal{Z}}$:
 \begin{equation}
  \nu^2_{\mathcal{Z}} = \mrm{Var}_{\mathcal{Z}}(\vec{G}_{\mathcal{Z}}) = \frac{1}{M}\sum_{j=1}^{M} \delta_{}^2(\vec{G}_{\mathcal{Z}},\vec{Z}_j)
 \label{eq:geomvar}
 \end{equation}

To shift a set of tangent space points to vary around a parametrized mean $\vec{G}_\phi$, parallel transport on $\mathcal{S}^+_D$ can be used \cite{amari_information_2016}:
\begin{equation}
  \Gamma_{\vec{G}_{\mathcal{Z}}\rightarrow\vec{G}_\phi}(\vec{S}) = \vec{E}^T\vec{S}\vec{E}~,~~\vec{E} = (\vec{G}_{\mathcal{Z}}^{-1}\vec{G}_\phi)^{\frac{1}{2}}
  \label{eq:paralleltransport}
  \end{equation}
While, parallel transport is generally defined for tangent space vectors $\vec{S}$ \cite{absil_optimization_2009}, on $\mathcal{S}^+_D$ the same operations also apply directly to points on the manifold (i.e., $\vec{Z} \in \mathcal{Z}$) \cite{yair_parallel_2019,brooks_riemannian_2019}.

\section{Domain-specific batch normalization on $\mathcal{S}^+_D$}
\vspace{-.5em}
In this section, we review relevant batch normalization (BN) \cite{ioffe_batch_2015} variants with a focus on $\mathcal{S}^+_D$.
We then present SPDMBN and show in a theoretical analysis that the running estimate converges to the true Fréchet mean under reasonable assumptions.
At last, we combine the idea of domain-specific batch normalization (DSBN) \cite{chang_domainspecific_2019} with SPDMBN to form a SPDDSMBN layer.
Table\,\ref{tab:acronymns} provides a brief overview of related and proposed methods.

\begin{table}  
  \caption{Overview and differences of relevant batch normalization algorithms. The last column, denoted normalization, sumarizes which statistics are used to normalize the batch data during training.}
  \label{tab:acronymns}
  \centering
  \footnotesize
  \begin{tabular}{lrrrr}
    \toprule
    Acronym & $\mathcal{S}_D^+$ & domain-specific & momentum $\gamma$ & normalization \\
\midrule
    MBN \cite{yong_momentum_2020} & no & no & adaptive  & running stats \\    
    SPDBN \cite{kobler_controlling_2022} & yes & no & fixed & running stats \\
    SPDMBN (algorithm \ref{algo:spdmbn}) \emph{proposed} & yes & no & adaptive  & running stats \\
    DSBN \cite{chang_domainspecific_2019} & no & yes & fixed & batch stats \\
    SPDDSMBN (\ref{eq:spddsmbn}) \emph{proposed} & yes & yes & adaptive  & running stats \\
    \bottomrule
    \end{tabular}
\end{table}

\paragraph{Batch normalization}
\label{sec:batchnorm}
Batch normalization (BN) \cite{ioffe_batch_2015} is a widely used training technqiue in deep learning as BN layers speed up convergence and improve generalization via smoothing of the engery landscape \cite{santurkar_how_2018,yong_momentum_2020}.
A standard BN layer applies slightly different transformations during training and testing to independent and identically distributed (iid) observations $\vec{x}_j \in \mathbb{R}^d$ within the $k$-th minibatch $\mathcal{B}_k$ of size $M$ drawn from a dataset $\mathcal{T}$.
During training, the data are normalized using the batch mean $\vec{b}_k$ and variance $\vec{s}^2_k$, and then scaled and shifted to have a parametrized mean $\vec{g}_\phi$ and variance $\boldsymbol{\sigma}^2_\phi$.
Internally, the layer updates running estimates of the dataset's statistics ($\vec{g}_k, \boldsymbol{\sigma}^2_k$) during each training step $k$; the updates are computed via exponential smoothing with momentum parameter $\gamma$.
During testing, the running estimates are used.

Using batch statistics to normalize data during training rather than running estimates introduces noise whose level depends on the batch size \cite{yong_momentum_2020}; smaller batch sizes raise the noise level.
The introduced noise regularizes the training process, which can help to escape poor local minima in initial learning but also lead to underfitting.
Momentum BN (MBN) \cite{yong_momentum_2020} allows small batch sizes while avoiding underfitting. Like batch renormalization \cite{ioffe_batch_2017}, MBN uses running estimates during training and testing.
The key difference is that MBN keeps two sets of running statistics; one for training and one for testing.
The latter are updated conventionally, while the former are updated with momentum parameter $\gamma_{train}(k)$ that decays over training steps $k$.
MBN can, therefore, quickly escape poor local minima during initial learning and avoid underfitting at later stages \cite{yong_momentum_2020}.
\vspace{-.5em}
\paragraph{Batch normalization on $\mathcal{S}^+_D$}
It is intractable to compute the Fréchet mean $\vec{G}_{\mathcal{B}_k}$ for each minibatch $\mathcal{B}_k = \{\vec{Z}_j \in \mathcal{S}^+_D\}_{j=1}^M$, as there is no efficient algorithm to solve (\ref{eq:geommean}).
\citet{brooks_riemannian_2019} proposed Riemannian Batch Normalization (RBN) as a tractable approximation.
RBN approximateley solves (\ref{eq:geommean}) by aborting the iterative Karcher flow algorithm after one iteration.
To transform $\vec{Z}_j \in \mathcal{B}_k$ with estimated mean $\vec{B}_k$ to vary around $\vec{G}_\phi$, parallel transport (\ref{eq:paralleltransport}) is used.
The RBN input output transformation is then expressed as
\begin{equation}
  \mrm{RBN}(\vec{Z}_j; \vec{G}_\phi, \gamma) = \Gamma_{\vec{B}_k\rightarrow\vec{G}_\phi}(\vec{Z}_j) = \vec{E}^T\vec{Z}_j\vec{E}~,~~\vec{E} = (\vec{B}_k^{-1}\vec{G}_\phi)^{\frac{1}{2}} ~, ~~~ \forall \vec{Z}_j\in\mathcal{B}_k
\label{eq:rbn}
\end{equation}
Using (\ref{eq:mean2points}), the running estimate of the dataset's Fréchet mean can be updated in closed form
\begin{equation}
  \vec{G}_k = \vec{G}_{k-1} \#_{\gamma} \vec{B}_k
  \label{eq:spdrunningmean}
\end{equation}
In \cite{kobler_controlling_2022} we proposed an extension to RBN, denoted SPD batch renormalization (SPDBN) that controls both Fréchet mean and variance.
Like batch renormalization \cite{ioffe_batch_2017}, SPDBN uses running estimates $\vec{G}_k$ and $\nu^2_k$ during training and testing.
To transform $\vec{Z}_j \in \mathcal{B}_k$ to vary around $\vec{G}_\phi$ with variance $\nu^2_\phi$, each observation is first transported to vary around the identity matrix $\vec{I}$, rescaled via computing matrix powers and finally transported to vary around $\vec{G}_\phi$. The sequence of operations can be expressed as
\begin{equation}
 \mrm{SPDBN}(\vec{Z}_j; \vec{G}_\phi, \nu^2_\phi, \varepsilon, \gamma) = \Gamma_{\vec{I}\rightarrow\vec{G}_\phi} \circ \Gamma_{\vec{G}_k\rightarrow\vec{I}}(\vec{Z}_j)^{\frac{\nu_\phi}{\nu_k + \varepsilon}}~, ~~~ \forall \vec{Z}_j\in\mathcal{B}_k
 \label{eq:spdbatchnorm}
\end{equation}
The standard backpropagation framework with extensions for structured matrices \cite{ionescu_matrix_2015} and manifold-constrained gradients \cite{absil_optimization_2009} can be used to propagate gradients through RBN and SPDBN layers and learn the parameters ($\vec{G}_\phi, \nu_\phi$).
\vspace{-.5em}
\paragraph{Momentum batch normalization on $\mathcal{S}^+_D$}
SPDBN \cite{kobler_controlling_2022} suffers from the same limitations as batch renormalization \cite{ioffe_batch_2017}.
Consequently, we propose to extend MBN \cite{yong_momentum_2020} to $\mathcal{S}^+_D$.
We list the pseudocode of our proposed extension, which we denote SPDMBN, in algorithm \ref{algo:spdmbn}.
SPDMBN uses approximations of batch-specific Fréchet means to update two sets of running estimates of the dataset's Fréchet mean.
As MBN \cite{yong_momentum_2020}, we decay $\gamma_{train}(k)$ with a clamped exponential decay schedule
\begin{equation}
  \gamma_{train}(k) = 1 - \gamma_{min}^{\frac{1}{K-1} \mrm{max}(K-k,0)} + \gamma_{min}
  \label{eq:momentumschedule}
\end{equation}
where $K$ defines the training step at which $\gamma_{min} \in [0,1]$ should be attained.

\begin{algorithm}[t]
  \footnotesize
  \SetAlgoLined
  \SetKwInOut{Input}{Input}\SetKwInOut{Output}{Output}
  \Input{batch $\mathcal{B}_k = \{\vec{Z}_j \in \mathcal{S}^+_D \}_{j = 1}^M$ at training step $k$, \\running mean $\bar{\vec{G}}_{k-1}$, $\bar{\vec{G}}_0 = \vec{I}$ and variance $\bar{\nu}^2_{k-1}$, $\bar{\nu}^2_0 = 1$ for training, \\
  running mean $\tilde{\vec{G}}_{k-1}$, $\tilde{\vec{G}}_0 = \vec{I}$ and variance $\tilde{\nu}^2_{k-1}$, $\tilde{\nu}^2_0 = 1$ for testing,\\
  learnable parameters $(\vec{G}_{\phi},\nu_{\phi})$, and momentum for training and testing $\gamma_{train}(k), \gamma \in [0,1]$}
  \Output{normalized batch $\{\tilde{\vec{Z}}_j = \mrm{SPDMBN}(\vec{Z}_j) \in \mathcal{S}^+_D ~|~ \vec{Z}_j\in\mathcal{B}_k \} $}
  \BlankLine
  \If{\texttt{training}}
  { 
    $\vec{B}_k =$ \texttt{karcher\_flow} ($\mathcal{B}_k$, \texttt{steps = 1})\tcp*[r]{approx. solve problem (\ref{eq:geommean})}

    $ \bar{\vec{G}}_k = \bar{\vec{G}}_{k-1} \#_{\gamma_{train}(k)} \vec{B}_k$\tcp*[r]{update running stats for training}

    $\bar{\nu}^2_k = (1 - \gamma_{train}(k)) \bar{\nu}^2_{k-1} + \gamma_{train}(k) \mrm{Var}_{\mathcal{B}_k}(\bar{\vec{G}}_k)$\\

    $ \tilde{\vec{G}}_k = \tilde{\vec{G}}_{k-1} \#_{\gamma} \vec{B}_k$\tcp*[r]{update running stats for testing}

    $\tilde{\nu}^2_k = (1 - \gamma) \tilde{\nu}^2_{k-1} + \gamma \mrm{Var}_{\mathcal{B}_k}(\tilde{\vec{G}}_k)$
  }
  $(\vec{G}_k, \nu^2_k) = (\bar{\vec{G}}_k, \bar{\nu}_k)$ \textbf{if} \texttt{\textit{training}} \textbf{else} $(\tilde{\vec{G}}_k, \tilde{\nu}^2_k)$
  
  $ \tilde{\vec{Z}}_j = \Gamma_{\vec{I}\rightarrow\vec{G}_\phi} \circ \Gamma_{\vec{G}_k\rightarrow\vec{I}}(\vec{Z}_j)^{\frac{\nu_\phi}{\nu_k + \varepsilon}} $ \tcp*[r]{use (\ref{eq:spdbatchnorm}) to whiten, rescale and rebias}

  \caption{SPD momentum batch normalization (SPDMBN)}\label{algo:spdmbn}
\end{algorithm}
\vspace{-.5em}
\paragraph{The running mean in SPDMBN converges to the Fréchet mean}
\label{par:spdmbn_converges_to_frechetmean}
Here, we consider models that apply a SPDMBN layer to latent representations generated by a feature extractor $f_\theta : \mathcal{X} \rightarrow \mathcal{S}^+_D$ with learnable parameters $\theta$. 

We define a dataset that contains the latent representations generated with feature set $\theta_k$ as $\mathcal{Z}_{\theta_k} = \{ f_{\theta_k} (\vec{x}) | \vec{x} \in \mathcal{T} \}$, and a minibatch of $M$ iid samples at training step $k$ as $\mathcal{B}_k$.
We denote the Fréchet mean of $\mathcal{Z}_{\theta_k}$ as $\vec{G}_{\theta_k}$, the estimated Fréchet mean, defined in (\ref{eq:spdrunningmean}), as $\vec{G}_k$, and the estimated batch mean as $\vec{B}_k$. Since the batches are drawn randomly, we consider the batch and running means as random variables.

We assume that the variance $\mrm{Var}_{\theta_{k}}(\vec{B}_{k-1}) = \mathbb{E}_{\vec{B}_{k-1}} \{ \delta^2(\vec{B}_{k-1}, \vec{G}_{\theta_{k}}) \}$ of the previous batch mean $\vec{B}_{k-1}$ with respect to the current Fréchet mean $\vec{G}_{\theta_k}$ is bounded by the current variance $\mrm{Var}_{\theta_{k}}(\vec{B}_{k})$ and the norm of the difference in the parameters
\begin{equation}
  \mrm{Var}_{\theta_k}(\vec{B}_{k-1}) \le (1 + || \theta_k - \theta_{k-1} ||) \mrm{Var}_{\theta_k}(\vec{B}_{k}) 
  \label{eq:assumption1}
\end{equation}
That is, across training steps $k$ the parameter updates are required to change the first and second order moments of the distribution of $\mathcal{Z}_{\theta_k}$ gradually so that the expected distance between $\vec{G}_{\theta_k}$ and the previous batch mean $\vec{B}_{k-1}$ is bounded.
We conjecture that this is the case for feature extractors $f_\theta$ that are smooth in the parameters and small learning rates, but leave the proof for future work.



\begin{proposition}[Error bound for $\vec{G}_k$]
  Consider the setting defined above, and assumption (\ref{eq:assumption1}) holds true. Then, the variance of the running mean $\mrm{Var}_{\theta_k}(\vec{G}_k)$ is bounded by  
  \begin{equation}
    \mrm{Var}_{\theta_k}(\vec{G}_k) \le \mrm{Var}_{\theta_k}(\vec{B}_k)
    \label{eq:spdbnboundvariance}
  \end{equation}
  over training steps $k$ if
  \begin{equation}
    || \theta_k - \theta_{k-1} || \le  \frac{1- \gamma^2}{(1-\gamma)^2} - 1
    \label{eq:spdbnbound}
  \end{equation}
  holds true.
  \label{prop:spdbnbound}
\end{proposition}
The proof is provided in appendix \ref{appendix:proof} of the supplementary material and relies on the proof of the geometric law of large numbers \cite{ho_recursive_2013}.

Proposition \ref{prop:spdbnbound} states that if (\ref{eq:assumption1}) and (\ref{eq:spdbnbound}) are met, the expected distance between the true Fréchet mean and the running mean is less or equal to the one of the batch mean.
Consequently, the introduced noise level of SPDBN (equation \ref{eq:spdbatchnorm}) and SPDMBN (algorithm \ref{algo:spdmbn}), which use $\vec{G}_k$ to normalize batches during training, is smaller or equal to RBN (equation \ref{eq:rbn}), which uses $\vec{B}_k$.

Since $\gamma$ controls the adaptation speed of $\vec{G}_k$, proposition \ref{prop:spdbnbound} also states that if $\gamma$ converges to zero (=no adaptation), the parameter updates are required to converge to zero as well (=no learning).
Hence, for a fixed $\gamma \in (0,1)$, as in the case of SPDBN (equation \ref{eq:spdbatchnorm}), proposition \ref{prop:spdbnbound} is fulfilled, if the learning rate for the parameters $\theta$ is chosen sufficiently small.
This can substantially slow down initial learning for standard choices of $\gamma$ (e.g., 0.1 or 0.01).
As a remedy, SPDMBN (algorithm \ref{algo:spdmbn}) uses an adaptive momentum parameter, which allows larger parameter updates during initial training steps.


If we consider a late stage of learning, and in particular assume that after a certain number of iterations $\kappa$ the parameters stay in a small ball with radius $\rho$ around 
$\theta^*$ (i.e., $|| \theta_k - \theta^* || \le \rho ~~~\forall~k > \kappa$)
and the feature extractor is $L$-smooth in the parameters (i.e., $\delta(f_\theta(\mathbf{x}), f_{\tilde{\theta}}(\mathbf{x})) \le L || \theta - \tilde{\theta} || ~ \forall \mathbf{x} \in \mathcal{T} ~,~ \forall \theta,\tilde{\theta}$)
then the distances are bounded $\delta(f_{\theta_k}(\mathbf{x}), f_{\theta^*}(\mathbf{x})) \le \rho L $.

\begin{remark}[Convergence of $\vec{G}_k$ for SPDMBN] {\normalfont If $\rho L$ is neglibile compared to the dataset's variance, then the Fréchet mean and variance can be considered fixed, and the theorem of large numbers on $\mathcal{S}^+_D$ [50] applies directly. That is, if the momentum parameter is decayed exponentially $\forall k > \kappa$ the running mean $\mathbf{G}_k$ converges to the Fréchet mean $\mathbf{G}_{\theta^*}$ in probability as $k \rightarrow \infty $.}
\label{remark:spdmbnconvergence}
\end{remark}

Taken together, Proposition \ref{prop:spdbnbound} and Remark \ref{remark:spdmbnconvergence} provide guidelines to update $\vec{G}_k$ in SPDMBN so that the introduced estimation error is bounded during initial fast learning (large $\gamma$) and decays towards zero in late learning (small $\gamma$).

\vspace{-.5em}
\paragraph{SPDMBN to learn tangent space mapping at Fréchet means}

Typical TSM models for classification \cite{Barachant2012} and regression \cite{sabbagh_manifold-regression_2019} first use (\ref{eq:log}) to project $\vec{Z} \in \mathcal{T} \subset \mathcal{S}_D^+$ to the tangent space at the Fréchet mean $\vec{G}_\mathcal{T}$, then use (\ref{eq:paralleltransport}) to transport the result to vary around $\vec{I}$, and finally extract elements in the upper triangular part\footnote{To preserve the norm, the off diagonal elements are scaled by $\sqrt{2}$.} to reduce feature redundancy.
The invertible mapping $\mathcal{P}_{\vec{G}_\mathcal{T}} : \mathcal{S}_D^+ \rightarrow \mathbb{R}^{D(D+1)/2}$ is expressed as:
\begin{equation}
  \mathcal{P}_{\vec{G}_\mathcal{T}}(\vec{Z}) = \mrm{upper} \circ \Gamma_{\vec{G}_\mathcal{T}\rightarrow\vec{I}} \circ \mathrm{Log}_{\vec{G}_\mathcal{T}}(\vec{Z}) = \mrm{upper}(\mathrm{log}(\vec{G}_\mathcal{T}^{-\frac{1}{2}}\vec{Z}\vec{G}_\mathcal{T}^{-\frac{1}{2}}))
  \label{eq:tsm}
\end{equation}

We propose to use a SPDMBN layer followed by a LogEig layer \cite{huang_riemannian_2017}  to compute a similar mapping $m_\phi$ (Figure\,\ref{fig:overview}a).
A LogEig layer simply computes the matrix logarithm and vectorizes the result so that the norm is preserved.
If the parametrized mean of SPDMBN is fixed to the identify matrix ($\vec{G}_\phi= \vec{I}$), the composition computes
\begin{align}
  m_\phi(\vec{Z}) & = \mrm{LogEig} \circ \mrm{SPDMBN} (\vec{Z}) = \mrm{upper} \circ \mathrm{log} \circ \Gamma_{\vec{G}_k\rightarrow\vec{I}}(\vec{Z})^{\frac{\nu_\phi}{\nu_k + \varepsilon}} \nonumber \\
   & = \mrm{upper} \left( \frac{\nu_\phi}{\nu_k + \varepsilon} \mathrm{log} \left( \vec{G}_k^{-\frac{1}{2}}\vec{Z}\vec{G}_k^{-\frac{1}{2}} \right)\right)
   \label{eq:tsmlayer}
\end{align}
where $(\vec{G}_k, \nu^2_k)$ are the estimated Fréchet mean and variance of the dataset $\mathcal{T}$ at training step $k$, and $\phi = \{ \nu_\phi \}$ the learnable parameters.
According to remark \ref{remark:spdmbnconvergence} $\vec{G}_k $ converges to $\vec{G}_{\mathcal{T}}$ and, in turn, $m_\phi$ to a scaled version of $\mathcal{P}_{\vec{G}_\mathcal{T}}$, since $\mrm{upper}$ is linear.

The mapping $m_\phi$ offers several advantageous properties.
First, the features are projected to a Euclidean vector space where standard layers can be applied and distances are cheap to compute.
Second, distances between the projected features locally approximate $\delta_{}$ and, therefore, inherit its invariance properties (e.g., affine mixing) \cite{pennec_riemannian_2006}.
This improves upon a LogEig layer \cite{huang_riemannian_2017} which projects features to the tangent space at the identity matrix.
As a result, distances between LogEig projected features correspond to distances measured with the log-Euclidean Riemannian metric (LERM) \cite{arsigny_geometric_2007} which is not invariant to affine mixing.
Third, controlling the Fréchet variance in (\ref{eq:tsmlayer}) empirically speeds up learning and improves generalization \cite{kobler_controlling_2022}.

\paragraph{Domain-specific batch normalization on $\mathcal{S}^+_D$}
Considering a multi-source UDA scenario, \citet{chang_domainspecific_2019} proposed a domain-specific BN (DSBN) layer which simply keeps multiple parallel BN layers and distributes observations according to the associated domains.
Formally, we consider minibatches $\mathcal{B}_k$ that form the union of $ N_{\mathcal{B}_k} \le |\mathcal{I}_d|$ domain-specific minibatches $\mathcal{B}_k^i$ drawn from distinct domains $i \in \mathcal{I}_{\mathcal{B}_k} \subseteq \mathcal{I}_d$.
As before, each $\mathcal{B}_k^i$ contains $j=1,...,M/N_{\mathcal{B}_k}$ iid observations $\vec{x}_{j}$.
A DSBN layer mapping $ \mathbb{R}^d\times \mathcal{I}_d \rightarrow \mathbb{R}^d$ can then be expressed as
\begin{equation}
  \mrm{DSBN}(\vec{x}_j, i) = \mrm{BN}_i(\vec{x}_j; \vec{g}_{\phi_i},  \vec{s}_{\phi_i}, \varepsilon, \gamma) ~, ~~~ \forall \vec{x}_j\in\mathcal{B}_k^i ~,~~~\forall i \in \mathcal{I}_{\mathcal{B}_k}
  \label{eq:dsbn}
\end{equation}
In practice, the batch size $M$ is typically fixed.
The particular choice is influenced by resource availability and the desired noise level introduced by minibatch based stochastic gradient descent.
A drawback of DSBN is that for a fixed batch size $M$ and an increasing number of source domains $N_{\mathcal{B}_k}$, the effective batch size declines for the BN layers within DSBN.
Since small batch sizes increase the noise level introduced by BN, increasing the number of domains per batch can lead to underfitting \cite{yong_momentum_2020}.
To alleviate this effect, we use the previously introduced SPDMBN layer. 
The proposed domain-specific BN layer on $\mathcal{S}_D^+$ is then formally defined as 
\begin{equation}
  \mrm{SPDDSMBN}(\vec{Z}_j, i) = \mrm{SPDMBN}_i(\vec{Z}_j; \vec{G}_{\phi_i}, \nu_{\phi_i}, \varepsilon, \gamma, \gamma_{train}(k)) ~, ~ \forall \vec{Z}_j\in\mathcal{B}_k^i \subset \mathcal{S}^+_D  ~,~\forall i \in \mathcal{I}_{\mathcal{B}_k}
  \label{eq:spddsmbn}
\end{equation}
The layer can be readily adapted to new domains, as new SPDMBN layers can be added on the fly.
If the entire data of a domain becomes available, the domain-specific Fréchet mean and variance can be estimated by solving (\ref{eq:geommean}), otherwise, the update rules in algorithm \ref{algo:spdmbn} can be used.

\section{SPDDSMBN to crack interpretable multi-source/-target UDA for EEG data}

With SPDDSMBN introduced in the previous section, we focus on a specific application domain, namely, multi-source/-target UDA for EEG-based BCIs and propose an intrinsically interpretable architecture which we denote TSMNet.
\vspace{-.5em}
\paragraph{Generative model of EEG}
EEG signals $\vec{x}(t) \in \mathbb{R}^P$ capture voltage fluctuations on $P$ channels.
An EEG record (=domain) is uniquely identified by a subject and session identifier.
After standard pre-processing steps, each domain $i$ contains $j = 1,...,M$ labeled observations with features $\vec{X}_{ij} \in \mathcal{X} \subset \mathbb{R}^{P\times T}$ where $T$ is the number of temporal samples.
Due to linearity of Maxwell's equations and Ohmic conductivity of tissue layers in the frequency ranges relevant for EEG \cite{nunez_electric_2006}, a domain-specific linear instantaneous mixture of sources model is a valid generative model:
\begin{equation}
  \vec{X}_{ij} = \vec{A}_{i} \vec{S}_{ij} + \vec{N}_{ij}
\end{equation}
where $\vec{S}_{ij} \in \mathbb{R}^{Q\times T}$ represents the activity of $Q$ latent sources, $\vec{A}_{i} \in \mathbb{R}^{P\times Q}$ a domain-specific mixing matrix and $\vec{N}_{ij} \in \mathbb{R}^{P\times T}$ additive noise. 
Both $\vec{A}_{i}$ and $\vec{S}_{ij}$ are unknown which demands making assumptions on $\vec{A}_{i}$ (e.g., anatomical prior knowledge \cite{Michel2004}) and/or $\vec{S}_{ij}$ (e.g., statistical independence \cite{hyvarinen_independent_2000}) to extract interesting sources.
\paragraph{Interpretable multi-source/-target UDA for EEG data}
As label information is available for the source domains, our goal is to identify discriminative oscillatory sources shared across domains.
Our approach relies on TSM models with linear classifiers \cite{Barachant2012}, as they are consistent \cite{sabbagh_manifold-regression_2019} and intrinsically interpretable \cite{kobler_interpretation_2021} estimators for generative models with log-linear relationships between the target $y_{ij}$ and variance $\mrm{Var}\{s_{ij}^{(k)}(t)\}$ of $k=1,...,K \le Q$ discriminative sources:
\begin{equation}
  y_{ij} = \sum_{k=1}^K b_k \mrm{log}\left( \mrm{Var}\{s_{ij}^{(k)}(t)\} \right)+ \varepsilon_{ij}
  \label{eq:encoding_model}
\end{equation}
where $b_k \in \mathbb{R}$ summarizes the coupling between the target $y_{ij}$ and the variance of the encoding source, and $\varepsilon_{ij}$ additive noise.
In \cite{kobler_interpretation_2021} we showed that the encoding sources' coupling and their patterns\footnote{Here, we use the entire dataset's Fréchet mean instead of the domain-specific ones to compute patterns for the average domain.} (columns of $\vec{A}_{i}$) can be recovered via solving a generalized eigenvalue problem between the Fréchet mean $\vec{G}_{\mathcal{T}_i}$ and classifier patterns \cite{haufe_on_2014} that were back projected to $\mathcal{S}_D^+$ with $\mathcal{P}_{\vec{G}_{\mathcal{T}_i}}^{-1}$.
The resulting eigenvectors are the patterns and the eigenvalues $\lambda_k$ reflect the relative source contribution $c_k$:
\begin{equation}
  c_k = \mrm{max}(\lambda_k,\lambda_k^{-1})~,~~~\lambda_k = \mrm{exp}(b_k/||\vec{b}||_2^2)
  \label{eq:pattern_contribution}
\end{equation}
To benefit from the intrinsic interpretability of TSM models, we constrain our hypothesis class $\mathcal{H}$ to functions $h : \mathcal{X}\times \mathcal{I}_d \rightarrow \mathcal{Y}$ that can be decomposed into a composition of a shared linear feature extractor with covariance pooling $f_{\theta} : \mathcal{X} \rightarrow \mathcal{S}_D^+$, domain-specific tangent space mapping $m_{\phi} : \mathcal{S}_D^+ \times \mathcal{I}_d \rightarrow \mathbb{R}^{D(D+1)/2}$, and a shared linear classifier $g_{\psi}: \mathbb{R}^{D(D+1)/2} \rightarrow \mathcal{Y}$ with parameters $\Theta = \{\theta, \phi, \psi\}$.
\vspace{-.5em}
\paragraph{TSMNet with SPDDSMBN}
Unlike previous approaches which learn $f_{\theta}, m_{\phi}, g_{\psi}$ sequentially \cite{zanini_transfer_2018,yair_parallel_2019,sabbagh_manifold-regression_2019,rodrigues_riemannian_2019}, we parametrize $h = g_{\psi} \circ m_{\phi} \circ f_{\theta}$ as a neural network and learn the entire model in an end-to-end fashion (Figure\,\ref{fig:overview}b).
Details of the proposed architecture, denoted TSMNet, are provided in appendix \ref{appendix:methods:tsmnet}.
In a nutshell, we parametrize $f_{\theta}$ as the composition of the first two linear convolutional layers of ShConvNet \cite{schirrmeister_deep_2017}, covariance pooling \cite{acharya_covariance_2018}, BiMap \cite{huang_riemannian_2017}, and ReEig \cite{huang_riemannian_2017} layers.
A BiMap layer applies a linear subspace projection, and a ReEig layer thresholds eigenvalues of symmetric matrices so that the output is SPD.
We used the default threshold ($10^{-4}$) and found that it was never active in the trained models. 
Hence, after training, $f_{\theta}$ fulfilled the hypothesis class constraints.
In order for $m_{\phi}$ to align the domain data and compute TSM, we use SPDDSMBN (\ref{eq:spddsmbn}) with shared parameters (i.e., $\vec{G}_{\phi_i} = \vec{G}_{\phi} = \vec{I}, \nu_{\phi_i} = \nu_{\phi}$) in (\ref{eq:tsmlayer}).
Finally, the classifier $g_{\psi}$ was parametrized as a linear layer with softmax activations. 
We use the standard-cross entropy loss as training objective, and optimized the parameters with the Riemannian ADAM optimizer \cite{becigneul_riemannian_2019}.

\section{Experiments with EEG data}
\label{sec:experiments}

In the following, we apply our method to classify target labels from short segments of EEG data.
We consider two BCI applications, namely, mental imagery \cite{pfurtscheller_motor_2001,Wolpaw2002} and mental workload estimation \cite{dehais_neuroergonomics_2020}.
Both applications have high potential to aid society in rehabilitation and healthcare \cite{donati_long-term_2016,novak_benchmarking_2018,fairclough_grand_2020} but have, currently, limited practical value because of poor generalization across sessions and subjects \cite{wei_2021_2022,roy_retrospective_2022}.
\vspace{-.5em}
\paragraph{Datasets and preprocessing}
\label{par:experiments:datasets}
The considered mental imagery datasets were BNCI2014001 \cite{Tangermann2012} (9 subjects/2 sessions/4 classes), BNCI2015001 \cite{Faller2012} (12/2-3/2), Lee2019 \cite{lee_eeg_2019} (54/2/2), Lehner2020 \cite{lehner_design_2020} (1/7/2), Stieger2021 \cite{stieger_continuous_2021} (62/4-8/4) and Hehnberger2021 \cite{hehenberger_long-term_2021} (1/26/4).
For mental workload estimation, we used a recent competition dataset \cite{hinss_eegdata_2021} (12/2/3).
A detailed description of the datasets is provided in appendix \ref{appendix:methods:datasets}.
Altogether, we analyzed a total of 603 sessions of 158 human subjects whose data was acquired in previous studies that obtained the subjects' informed consent and the right to share anonymized data.\\
The python packages moabb \cite{jayaram_moabb_2018} and mne \cite{gramfort_meg_2013} were used to preprocess the datasets.
The applied steps comprise resampling the EEG signals to 250/256 Hz, applying temporal filters to extract oscillatory EEG activity in the 4 to 36 Hz range (spectrally resolved if required by a method) and finally extract short segments ($\le 3 s$) associated to a class label (details provided in appendix \ref{appendix:methods:preprocessing}).
\vspace{-.5em}
\paragraph{Evaluation}
\label{par:experiments:evaluation}
We evaluated TSMNet against several baseline methods implementing direct transfer or multi-source (-target) UDA strategies.
They can be broadly categorized as component based \cite{kai_keng_ang_filter_2008,hehenberger_long-term_2021}, Riemannian geometry aware \cite{Barachant2012,kobler_interpretation_2021,rodrigues_riemannian_2019,yair_domain_2020} or deep learning \cite{schirrmeister_deep_2017,lawhern_eegnet_2018,ozdenizci_learning_2020}. 
All models were fit and evaluated with a randomized leave 5\% of the sessions (inter-session TL) or subjects (inter-subject TL) out cross-validation (CV) scheme.
For inter-session TL, the models were only provided with data of the associated subject.
When required, inner train/test splits (neural nets) or CV (shallow methods) were used to optimize hyper parameters (e.g., early stopping, regularization parameters).
The dataset Hehenberger2021 was used to fit the hyper parameters of TSMNet, and is, therefore, omitted in the presented results.
Balanced accuracy (i.e., the average recall across classes) was used as scoring metric.
As the discriminability of the data varies considerably across subjects, we decided to report the results in the figures relative to the score of a SoA domain-specific Riemannian geometry aware method \cite{kobler_interpretation_2021}, which was fitted and evaluated in a 80\%/20\% chronological train/test split (for details see appendix \ref{appendix:methods:dsreference}).

\vspace{-.5em}
\paragraph{Soft- and hardware}
\label{par:experiments:softhardware}
We either used publicly available python code or implemented the methods in python using the packages torch \cite{paszke_pytorch_2019}, scikit-learn \cite{pedregosa_scikitlearn_2011}, skorch \cite{tietz_skorch_2017}, geoopt \cite{kochurov_geoopt_2020}, mne \cite{gramfort_meg_2013}, pyriemann \cite{alexandre_pyriemann_2022}, pymanopt \cite{townsend_pymanopt_2016}.
We ran the experiments on standard computation PCs equipped with 32 core CPUs with 128 GB of RAM and used up to 1 GPU (24 GRAM).
Depending on the dataset size, fitting and evaluating TSMNet varied from a few seconds to minutes.  

\subsection{Mental imagery}
\begin{figure}
  \centering
  \includegraphics[width=\textwidth]{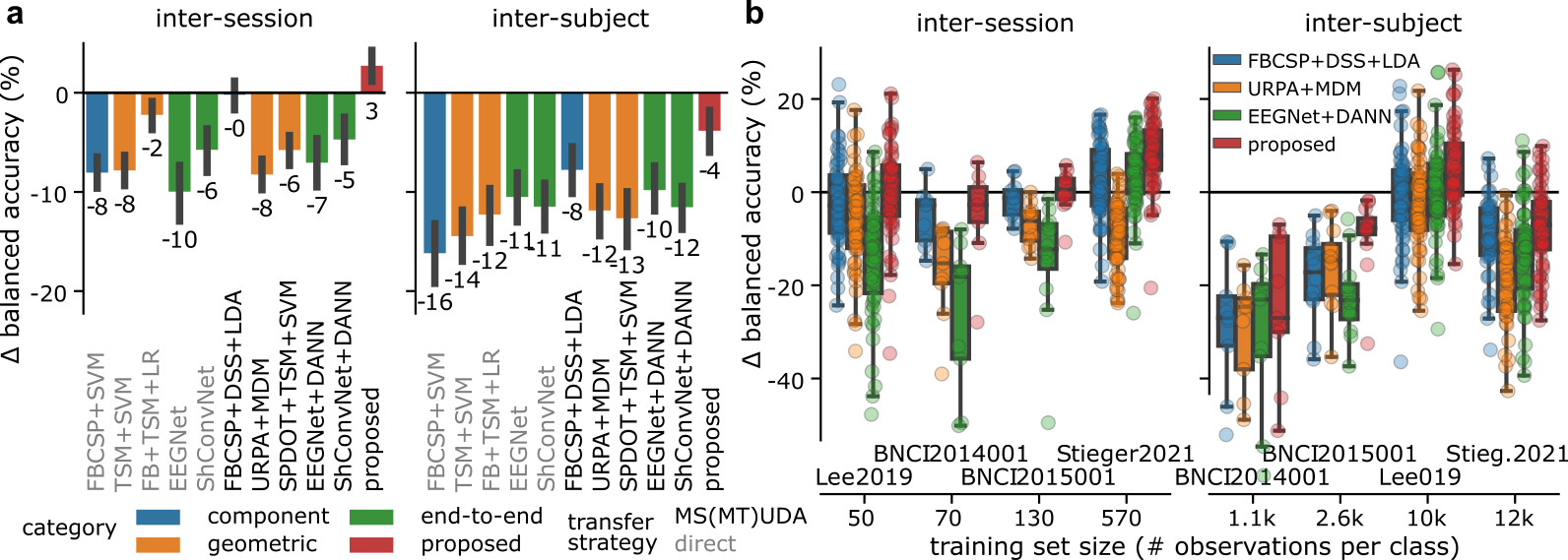}
  \caption{Mental imagery results (5 datasets). 
  BCI test set score (balanced accuracy) for inter-session/-subject transfer learning methods relative to a SoA domain-specific reference model (80\%/20\% chronological train/test split; for details see appendix \ref{appendix:methods:dsreference}). 
  \textbf{a}, Barplots summarize the grand average (573 sessions, 138 subjects) results.
  Errorbars indicate bootstrapped (1e3 repetitions) 95\% confidence intervals (over subjects).
  \textbf{b}, Box and scatter plots summarize the dataset-specific results for selected methods from each category.
  Datasets are ordered according to the training set size.
  Each dot summarizes the score for one subject.
  Lehner2021 is not displayed as it contains only 1 subject.
  }\label{fig:res-eegmi}
\end{figure}

\begin{figure}
  \centering
  \includegraphics[width=\textwidth]{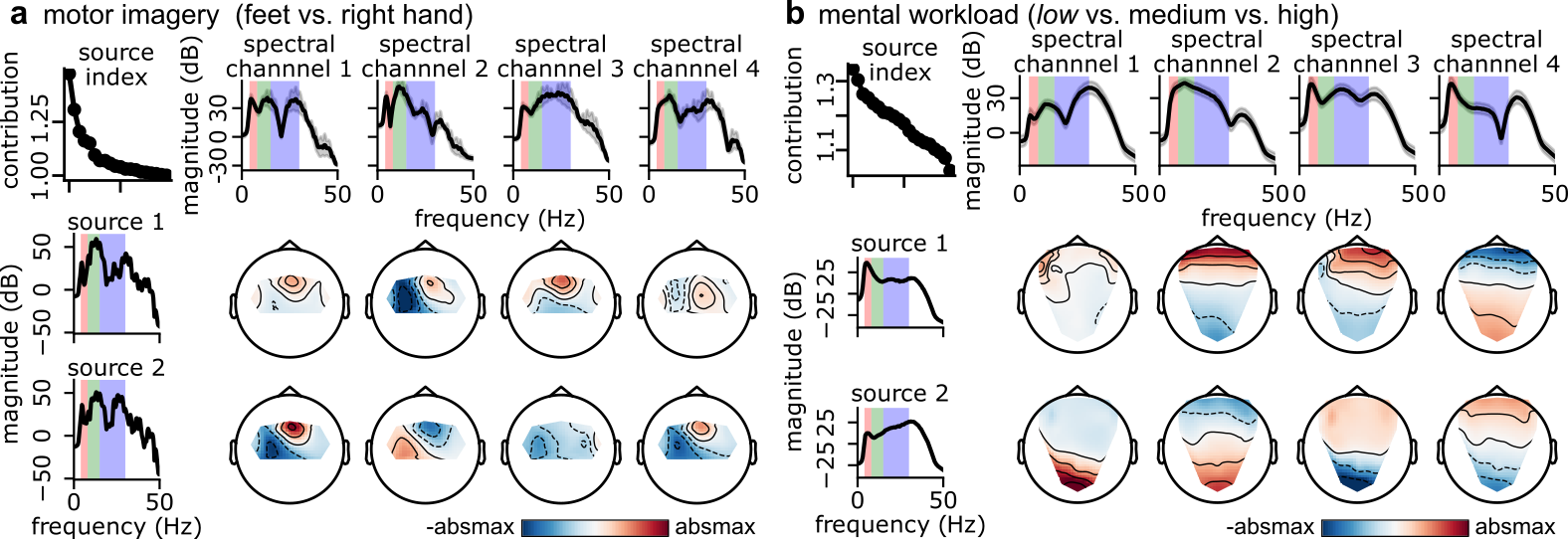}
  \caption{Model interpretation results. 
  Patterns extracted from a TSMNet.
  \textbf{a}, Motor imagery dataset (BNCI2015001, inter-subject TL).
  The top, left panel lists the contribution, defined in (\ref{eq:pattern_contribution}), for each extracted source $k=1,...,20$ (x-axis) to the target class.
  Panels in the left column summarize the spectral patterns of extracted sources.
  For visualization purposes, only the 2 most discriminative sources are displayed. 
  Panels in the top row summarize the frequency profile of each spectral channel (output of 4 temporal convolution layers in $f_\theta$).
  Topographic plots summarize how the source activity is projected to regions covered by the EEG channels (rows correspond to the source index; columns to spectral channels).
  EEG channels at darker blue or red areas capture more source activity and are, therefore, more discriminative.
  \textbf{b}, As in \textbf{a} for the mental workload estimation dataset and class \emph{low}.
  }\label{fig:res-patterns}
\end{figure}

\vspace{-.5em}
\paragraph{TSMNet closes the gap to domain-specific methods}
Figure\,\ref{fig:res-eegmi} summarizes the mental imagery results.
It displays test set scores of the considered TL methods relative to the score of a SoA domain-specific reference method.
Combining the results of all subjects across datasets (Figure\,\ref{fig:res-eegmi}a), it becomes apparent that TSMNet is the only method that can significantly reduce the gap to the reference method (inter-subject) or even exceed its performance (inter-session).
Figure\,\ref{fig:res-eegmi}b displays the results resolved across datasets (for details see appendix \ref{appendix:results:eeg}).
We make two important observations.
First, concerning inter-session TL, TSMNet meets or exceeds the score of the reference method consistently across datasets. 
Second, concerning inter-subject TL, we found that all considered methods tend to reduce the performance gap as the dataset size (\# subjects) increases, and that TSMNet is consistently the top or among the top methods.
As a fitted TSMNet corresponds to a typical TSM model with a linear classifier, we can transform the fitted parameters into interpretable patterns \cite{kobler_interpretation_2021}.
Figure\,\ref{fig:res-patterns}a displays extracted patterns for the BNCI2015001 dataset (inter-subject TL).
It is clearly visible that TSMNet infers the target label from neurophysiologically plausible sources (rows in Figure\,\ref{fig:res-patterns}a).
As expected \cite{pfurtscheller_event-related_1999}, the source with highest contribution has spectral peaks in the alpha and beta bands, and originates in contralateral and central sensorimotor cortex.

\vspace{-.5em}
\paragraph{DSBN on $\mathcal{S}_D^+$ drives the success of TSMNet}
Since TSMNet combines several advances, we present the results of an ablation study in Table\,\ref{tab:ablation}.
It summarizes the grand average inter-session TL test scores relative to TSMNet with SPDDSMBN for n = 138 subjects.
We observed three significant effects.
The largest effect can be attributed to $\mathcal{S}_D^+$, as we observed the largest performance decline if the architecture would be modified\footnote{We replaced the covariance pooling, BiMap, ReEig, SPD(DS)MBN, LogEig layers with variance pooling, elementwise log activations followed by (DS)MBN. Note that the resulting architecture is similar to ShConvNet.} to omit the SPD manifold (4.5\% with DSBN, 3\% w/o DSBN). 
The performance gain comes at the cost of a 2.6x longer time to fit the parameters.
The second largest effect can be attributed to DSBN; without DSBN the performance dropped by 3.9\% (with $\mathcal{S}_D^+$) and 2.4\% (w/o $\mathcal{S}_D^+$).
The smallest, yet significant effect can be attributed to SPDMBN.

\begin{table}  
  \caption{Ablation study.
  Grand average (5 mental imagery datasets, 138 subjects, inter-session TL) score for the test data relative to the proposed method, and training fit time (50 epochs). 
  Standard-deviation is used to report the variability across subjects.
  Permutation t-tests (1e4 perms, df=137, 4 tests with t-max adjustment) were used to identify significant effects.}
  \label{tab:ablation}
  \centering
  \footnotesize
  \begin{tabular}{lllrrr}
    \toprule
    & &                     & \multicolumn{2}{c}{ $\Delta$ balanced accuracy (\%)} &                               fit time (s) \\
    & &                     &                                 mean (std) & t-val (p-val) & mean (std) \\
    $\mathcal{S}_D^+$ & DSBN &BN method &                                   &                                            &                                            \\
\midrule
    yes & yes    & SPDMBN (algo. \ref{algo:spdmbn}) \emph{(proposed)} &             - &     - &  16.9\phantom{0}(\phantom{0}1.0) \\
        & yes    & SPDBN \cite{kobler_controlling_2022} &             -1.6\phantom{0}(\phantom{0}2.2) &   -7.8\phantom{0}(0.0001) &  20.3\phantom{0}(\phantom{0}1.6) \\
        & no     & SPDMBN (algo. \ref{algo:spdmbn}) &             -3.9\phantom{0}(\phantom{0}4.4) &  -10.7\phantom{0}(0.0001) &  11.3\phantom{0}(\phantom{0}0.5) \\   
    no  & yes    & MBN \cite{yong_momentum_2020} &             -4.5\phantom{0}(\phantom{0}3.8) &  -10.1\phantom{0}(0.0001) &  \phantom{0}6.6\phantom{0}(\phantom{0}0.2) \\     
        & no     & MBN \cite{yong_momentum_2020} &             -6.9\phantom{0}(\phantom{0}4.8) &  -13.4\phantom{0}(0.0001) &  \phantom{0}4.4\phantom{0}(\phantom{0}0.1) \\

    \bottomrule
    \end{tabular}
\end{table}
\vspace{-.5em}
\subsection{Mental workload estimation}
\vspace{-.5em}
Compared to the baseline methods, TSMNet obtained the highest average scores of 54.7\% (7.3\%) and 52.4\% (8.8\%) in inter-session and -subject TL (for details see appendix \ref{appendix:results:eeg}).
Interestingly, the inter-session TL score of TSMNet matches the score (54.3\%) of the winning method in last year's competition \cite{roy_retrospective_2022}. 
To shed light on the sources utilized by TSMNet, we show patterns for a fitted model in Figure\,\ref{fig:res-patterns}b.
For the low mental workload class, the top contributing source's activity peaked in the theta band and originated in pre-frontal areas.
The second source's activity originated in occipital cortex with non-focal spectral profile.
Our results agree with the findings of previous research, as both areas and the theta band have been implicated in mind wandering and effort withdrawal \cite{dehais_neuroergonomics_2020}.   
\vspace{-.5em}
\section{Discussion}
\label{sec:discussion}
\vspace{-.5em}
In this contribution, we proposed a machine learning framework around (domain-specific) momentum batch normalization on $\mathcal{S}_D^+$ to learn tangent space mapping (TSM) and feature extractors in an end-to-end fashion.
In a theoretical analysis, we provided error bounds for the running estimate of the Fréchet mean as well as convergence guarantees under reasonable assumptions.
We then applied the framework, to a multi-source multi-target unsupervised domain adaptation problem, namely, inter-session and -subject transfer learning for EEG data and obtained or attained state-of-the art performance with a simple, intrinsically interpretable model, denoted TSMNet, in a total of 6 diverse BCI datasets (138 human subjects, 573 sessions).
In the case of mental imagery, we found that TSMNet significantly reduced (inter-subject TL) or even exceeded (inter-session TL) the performance gap to a SoA domain-specific method.

Although our framework could be readily extended to online UDA for unseen target domains, we limited this study to offline evaluations and leave actual BCI studies to future work.
A limitation of our framework, and also any other method that involves eigen decomposition, is the computational complexity, which limits its application to high-dimensional SPD features (e.g., fMRI connectivity matrices with fine spatial granularity).
Altogether, the presented results demonstrate the utility of our framework and in particular TSMNet as it not only achieves highly competitive results but is also intrinsically interpretable.
While we do not foresee any immediate negative societal impacts, we provide direct contributions towards the scalability and acceptability of EEG-based healthcare \cite{Schomer2011,Wolpaw2002} and consumer \cite{fairclough_grand_2020,dehais_neuroergonomics_2020} technologies.
We expect future works to evaluate the impact of the proposed methods in clinical applications of EEG like sleep staging \cite{campbell_eeg_2009,banville_uncovering_2020}, seizure \cite{acharya_automated_2013} or pathology detection \cite{nuwer_routine_2005,gemein_machine-learning-based_2020}. 

\begin{ack}
  This work was supported by the JSPS KAKENHI (Grants-in-Aid for Scientific Research) grants 20H04249, 20H04208, 21H03516 and 21K12055.


\end{ack}

\bibliography{references}

\begin{thebibliography}{85}
\providecommand{\natexlab}[1]{#1}
\providecommand{\url}[1]{\texttt{#1}}
\expandafter\ifx\csname urlstyle\endcsname\relax
  \providecommand{\doi}[1]{doi: #1}\else
  \providecommand{\doi}{doi: \begingroup \urlstyle{rm}\Url}\fi

\bibitem[Schomer and Lopes~da Silva(2018)]{Schomer2011}
Donald~L. Schomer and Fernando~H. Lopes~da Silva.
\newblock \emph{Niedermeyer's {Electroencephalography}: basic principles,
  clinical applications, and related fields.}
\newblock Lippincott Williams \& Wilkins, 7 edition, 2018.
\newblock ISBN 978-0-19-022848-4.
\newblock \doi{10.1212/WNL.0b013e31822f0490}.

\bibitem[Pfurtscheller and Lopes(1999)]{pfurtscheller_event-related_1999}
G~Pfurtscheller and F~H Lopes.
\newblock Event-related {EEG} / {MEG} synchronization and.
\newblock \emph{Clinical Neurophysiology}, 110:\penalty0 10576479, 1999.

\bibitem[Faller et~al.(2019)Faller, Cummings, Saproo, and
  Sajda]{faller_regulation_2019}
Josef Faller, Jennifer Cummings, Sameer Saproo, and Paul Sajda.
\newblock Regulation of arousal via online neurofeedback improves human
  performance in a demanding sensory-motor task.
\newblock \emph{Proceedings of the National Academy of Sciences}, 116\penalty0
  (13):\penalty0 6482--6490, 2019.
\newblock \doi{10.1073/pnas.1817207116}.

\bibitem[Zhang et~al.(2021)Zhang, Wu, Toll, Naparstek, Maron-Katz, Watts,
  Gordon, Jeong, Astolfi, Shpigel, Longwell, Sarhadi, El-Said, Li, Cooper,
  Chin-Fatt, Arns, Goodkind, Trivedi, Marmar, and
  Etkin]{zhang_identification_2021}
Yu~Zhang, Wei Wu, Russell~T. Toll, Sharon Naparstek, Adi Maron-Katz, Mallissa
  Watts, Joseph Gordon, Jisoo Jeong, Laura Astolfi, Emmanuel Shpigel, Parker
  Longwell, Kamron Sarhadi, Dawlat El-Said, Yuanqing Li, Crystal Cooper,
  Cherise Chin-Fatt, Martijn Arns, Madeleine~S. Goodkind, Madhukar~H. Trivedi,
  Charles~R. Marmar, and Amit Etkin.
\newblock Identification of psychiatric disorder subtypes from functional
  connectivity patterns in resting-state electroencephalography.
\newblock \emph{Nature Biomedical Engineering}, 5\penalty0 (4):\penalty0
  309--323, 2021.
\newblock \doi{10.1038/s41551-020-00614-8}.

\bibitem[Wolpaw et~al.(2002)Wolpaw, Birbaumer, McFarland, Pfurtscheller, and
  Vaughan]{Wolpaw2002}
Jonathan~R Wolpaw, Niels Birbaumer, Dennis~J McFarland, Gert Pfurtscheller, and
  Theresa~M Vaughan.
\newblock Brain-computer interfaces for communication and control.
\newblock \emph{Clinical neurophysiology : official journal of the
  International Federation of Clinical Neurophysiology}, 113:\penalty0
  767--791, 2002.
\newblock \doi{doi: 10.1016/s1388-2457(02)00057-3}.

\bibitem[Ben-David et~al.(2010)Ben-David, Blitzer, Crammer, Kulesza, Pereira,
  and Vaughan]{ben-david_theory_2010}
Shai Ben-David, John Blitzer, Koby Crammer, Alex Kulesza, Fernando Pereira, and
  Jennifer~Wortman Vaughan.
\newblock A theory of learning from different domains.
\newblock \emph{Machine Learning}, 79\penalty0 (1-2):\penalty0 151--175, 2010.
\newblock \doi{10.1007/s10994-009-5152-4}.

\bibitem[Hoffman et~al.(2018)Hoffman, Mohri, and
  Zhang]{hoffmann_algorithms_2018}
Judy Hoffman, Mehryar Mohri, and Ningshan Zhang.
\newblock Algorithms and theory for multiple-source adaptation.
\newblock In S.~Bengio, H.~Wallach, H.~Larochelle, K.~Grauman, N.~Cesa-Bianchi,
  and R.~Garnett, editors, \emph{Advances in Neural Information Processing
  Systems}, volume~31. Curran Associates, Inc., 2018.

\bibitem[Wu et~al.(2020)Wu, Xu, and Lu]{wu_transfer_2020}
Dongrui Wu, Yifan Xu, and Bao-Liang Lu.
\newblock Transfer {Learning} for {EEG}-{Based} {Brain}-{Computer}
  {Interfaces}: {A} {Review} of {Progress} {Made} {Since} 2016.
\newblock \emph{IEEE Transactions on Cognitive and Developmental Systems},
  pages 1--1, 2020.
\newblock \doi{10.1109/TCDS.2020.3007453}.

\bibitem[Neuper et~al.(2005)Neuper, Scherer, Reiner, and
  Pfurtscheller]{neuper_imagery_2005}
Christa Neuper, Reinhold Scherer, Miriam Reiner, and Gert Pfurtscheller.
\newblock Imagery of motor actions: {Differential} effects of kinesthetic and
  visual–motor mode of imagery in single-trial {EEG}.
\newblock \emph{Cognitive Brain Research}, 25\penalty0 (3):\penalty0 668--677,
  2005.
\newblock \doi{10.1016/j.cogbrainres.2005.08.014}.

\bibitem[Lotte et~al.(2018)Lotte, Bougrain, Cichocki, Clerc, Congedo,
  Rakotomamonjy, and Yger]{lotte_review_2018}
F~Lotte, L~Bougrain, A~Cichocki, M~Clerc, M~Congedo, A~Rakotomamonjy, and
  F~Yger.
\newblock A review of classification algorithms for {EEG}-based
  brain–computer interfaces: a 10 year update.
\newblock \emph{Journal of Neural Engineering}, 15\penalty0 (3):\penalty0
  031005, 2018.
\newblock \doi{10.1088/1741-2552/aab2f2}.

\bibitem[Statthaler et~al.(2017)Statthaler, Schwarz, Steyrl, Kobler, Höller,
  Brandstetter, Hehenberger, Bigga, and Müller-Putz]{Statthaler2017}
Karina Statthaler, Andreas Schwarz, David Steyrl, Reinmar~J. Kobler,
  Maria~Katharina Höller, Julia Brandstetter, Lea Hehenberger, Marvin Bigga,
  and Gernot Müller-Putz.
\newblock Cybathlon experiences of the {Graz} {BCI} racing team {Mirage91} in
  the brain-computer interface discipline.
\newblock \emph{Journal of NeuroEngineering and Rehabilitation}, 14\penalty0
  (1):\penalty0 129, 2017.
\newblock \doi{10.1186/s12984-017-0344-9}.

\bibitem[Barachant et~al.(2012)Barachant, Bonnet, Congedo, and
  Jutten]{Barachant2012}
Alexandre Barachant, Stéphane Bonnet, Marco Congedo, and Christian Jutten.
\newblock Multiclass brain-computer interface classification by {Riemannian}
  geometry.
\newblock \emph{IEEE transactions on bio-medical engineering}, 59\penalty0
  (4):\penalty0 920--928, 2012.
\newblock \doi{10.1109/TBME.2011.2172210}.

\bibitem[Sabbagh et~al.(2019)Sabbagh, Ablin, Varoquaux, Gramfort, and
  Engemann]{sabbagh_manifold-regression_2019}
David Sabbagh, Pierre Ablin, Gaël Varoquaux, Alexandre Gramfort, and Denis~A
  Engemann.
\newblock Manifold-regression to predict from {MEG}/{EEG} brain signals without
  source modeling.
\newblock In \emph{Advances in {Neural} {Information} {Processing} {Systems}},
  pages 7323--7334, 2019.

\bibitem[Jayaram and Barachant(2018)]{jayaram_moabb_2018}
Vinay Jayaram and Alexandre Barachant.
\newblock {MOABB}: trustworthy algorithm benchmarking for {BCIs}.
\newblock \emph{Journal of Neural Engineering}, 15\penalty0 (6):\penalty0
  066011, 2018.
\newblock \doi{10.1088/1741-2552/aadea0}.

\bibitem[Roy et~al.(2022)Roy, Hinss, Darmet, Ladouce, Jahanpour, Somon, Xu,
  Drougard, Dehais, and Lotte]{roy_retrospective_2022}
Raphaëlle~N. Roy, Marcel~F. Hinss, Ludovic Darmet, Simon Ladouce, Emilie~S.
  Jahanpour, Bertille Somon, Xiaoqi Xu, Nicolas Drougard, Frédéric Dehais,
  and Fabien Lotte.
\newblock Retrospective on the {First} {Passive} {Brain}-{Computer} {Interface}
  {Competition} on {Cross}-{Session} {Workload} {Estimation}.
\newblock \emph{Frontiers in Neuroergonomics}, 3:\penalty0 838342, 2022.
\newblock \doi{10.3389/fnrgo.2022.838342}.

\bibitem[Congedo et~al.(2017)Congedo, Barachant, and
  Bhatia]{congedo_riemannian_2017}
Marco Congedo, Alexandre Barachant, and Rajendra Bhatia.
\newblock Riemannian geometry for {EEG}-based brain-computer interfaces; a
  primer and a review.
\newblock \emph{Brain-Computer Interfaces}, 4\penalty0 (3):\penalty0 155--174,
  2017.
\newblock \doi{10.1080/2326263X.2017.1297192}.

\bibitem[Kobler et~al.(2021)Kobler, Hirayama, Hehenberger, Lopes-Dias,
  Müller-Putz, and Kawanabe]{kobler_interpretation_2021}
Reinmar~J. Kobler, Jun-Ichiro Hirayama, Lea Hehenberger, Catarina Lopes-Dias,
  Gernot Müller-Putz, and Motoaki Kawanabe.
\newblock On the interpretation of linear {Riemannian} tangent space model
  parameters in {M}/{EEG}.
\newblock In \emph{Proceedings of the 43rd {Annual} {International}
  {Conference} of the {IEEE} {Engineering} in {Medicine} and {Biology}
  {Society} ({EMBC})}. IEEE, 2021.
\newblock \doi{10.1109/EMBC46164.2021.9630144}.

\bibitem[Fairclough and Lotte(2020)]{fairclough_grand_2020}
Stephen~H. Fairclough and Fabien Lotte.
\newblock Grand {Challenges} in {Neurotechnology} and {System}
  {Neuroergonomics}.
\newblock \emph{Frontiers in Neuroergonomics}, 1:\penalty0 602504, 2020.
\newblock \doi{10.3389/fnrgo.2020.602504}.

\bibitem[Wei et~al.(2022)Wei, Faisal, Grosse-Wentrup, Gramfort, Chevallier,
  Jayaram, Jeunet, Bakas, Ludwig, Barmpas, Bahri, Panagakis, Laskaris, Adamos,
  Zafeiriou, Duong, Gordon, Lawhern, Śliwowski, Rouanne, and
  Tempczyk]{wei_2021_2022}
Xiaoxi Wei, A.~Aldo Faisal, Moritz Grosse-Wentrup, Alexandre Gramfort, Sylvain
  Chevallier, Vinay Jayaram, Camille Jeunet, Stylianos Bakas, Siegfried Ludwig,
  Konstantinos Barmpas, Mehdi Bahri, Yannis Panagakis, Nikolaos Laskaris,
  Dimitrios~A. Adamos, Stefanos Zafeiriou, William~C. Duong, Stephen~M. Gordon,
  Vernon~J. Lawhern, Maciej Śliwowski, Vincent Rouanne, and Piotr Tempczyk.
\newblock 2021 {BEETL} {Competition}: {Advancing} {Transfer} {Learning} for
  {Subject} {Independence} \& {Heterogenous} {EEG} {Data} {Sets}.
\newblock \emph{arXiv:2202.12950 [cs, eess]}, 2022.

\bibitem[Jayaram et~al.(2016)Jayaram, Alamgir, Altun, Scholkopf, and
  Grosse-Wentrup]{jayaram_transfer_2016}
Vinay Jayaram, Morteza Alamgir, Yasemin Altun, Bernhard Scholkopf, and Moritz
  Grosse-Wentrup.
\newblock Transfer {Learning} in {Brain}-{Computer} {Interfaces}.
\newblock \emph{IEEE Computational Intelligence Magazine}, 11\penalty0
  (1):\penalty0 20--31, 2016.
\newblock \doi{10.1109/MCI.2015.2501545}.

\bibitem[Fazli et~al.(2009)Fazli, Popescu, Danóczy, Blankertz, Müller, and
  Grozea]{Fazli2009}
Siamac Fazli, Florin Popescu, Márton Danóczy, Benjamin Blankertz,
  Klaus-Robert Müller, and Cristian Grozea.
\newblock Subject-independent mental state classification in single trials.
\newblock \emph{Neural Networks}, 22\penalty0 (9):\penalty0 1305--1312, 2009.
\newblock \doi{10.1016/j.neunet.2009.06.003}.

\bibitem[Samek et~al.(2014)Samek, Kawanabe, and Müller]{Samek2014}
Wojciech Samek, Motoaki Kawanabe, and Klaus~Robert Müller.
\newblock Divergence-based framework for common spatial patterns algorithms.
\newblock \emph{IEEE Reviews in Biomedical Engineering}, 7:\penalty0 50--72,
  2014.
\newblock \doi{10.1109/RBME.2013.2290621}.

\bibitem[Kwon et~al.(2020)Kwon, Lee, Guan, and
  Lee]{kwon_subject-independent_2020}
O-Yeon Kwon, Min-Ho Lee, Cuntai Guan, and Seong-Whan Lee.
\newblock Subject-{Independent} {Brain}–{Computer} {Interfaces} {Based} on
  {Deep} {Convolutional} {Neural} {Networks}.
\newblock \emph{IEEE Transactions on Neural Networks and Learning Systems},
  31\penalty0 (10):\penalty0 3839--3852, 2020.
\newblock \doi{10.1109/TNNLS.2019.2946869}.

\bibitem[Zhao et~al.(2020)Zhao, Li, Reed, Xu, and
  Keutzer]{zhao_multi-source_2020}
Sicheng Zhao, Bo~Li, Colorado Reed, Pengfei Xu, and Kurt Keutzer.
\newblock Multi-source {Domain} {Adaptation} in the {Deep} {Learning} {Era}:
  {A} {Systematic} {Survey}, 2020.

\bibitem[Ozdenizci et~al.(2020)Ozdenizci, Wang, Koike-Akino, and
  Erdogmus]{ozdenizci_learning_2020}
Ozan Ozdenizci, Ye~Wang, Toshiaki Koike-Akino, and Deniz Erdogmus.
\newblock Learning {Invariant} {Representations} {From} {EEG} via {Adversarial}
  {Inference}.
\newblock \emph{IEEE Access}, 8:\penalty0 27074--27085, 2020.
\newblock \doi{10.1109/ACCESS.2020.2971600}.

\bibitem[Xu et~al.(2021)Xu, Ma, Meng, Xu, Jung, and Ming]{xu_improving_2021}
Lichao Xu, Zhen Ma, Jiayuan Meng, Minpeng Xu, Tzyy-Ping Jung, and Dong Ming.
\newblock Improving {Transfer} {Performance} of {Deep} {Learning} with
  {Adaptive} {Batch} {Normalization} for {Brain}-computer {Interfaces}
  $^{\textrm{*}}$.
\newblock In \emph{2021 43rd {Annual} {International} {Conference} of the
  {IEEE} {Engineering} in {Medicine} \& {Biology} {Society} ({EMBC})}, pages
  5800--5803, Mexico, 2021. IEEE.
\newblock ISBN 978-1-72811-179-7.
\newblock \doi{10.1109/EMBC46164.2021.9629529}.

\bibitem[Xu et~al.(2020)Xu, Xu, Ke, An, Liu, and Ming]{xu_crossdataset_2020}
Lichao Xu, Minpeng Xu, Yufeng Ke, Xingwei An, Shuang Liu, and Dong Ming.
\newblock Cross-{Dataset} {Variability} {Problem} in {EEG} {Decoding} {With}
  {Deep} {Learning}.
\newblock \emph{Frontiers in Human Neuroscience}, 14:\penalty0 103, 2020.
\newblock \doi{10.3389/fnhum.2020.00103}.

\bibitem[Mane et~al.(2021)Mane, Chew, Chua, Ang, Robinson, Vinod, Lee, and
  Guan]{mane_fbcnet_2021}
Ravikiran Mane, Effie Chew, Karen Chua, Kai~Keng Ang, Neethu Robinson, A.~P.
  Vinod, Seong-Whan Lee, and Cuntai Guan.
\newblock {FBCNet}: {A} {Multi}-view {Convolutional} {Neural} {Network} for
  {Brain}-{Computer} {Interface}.
\newblock \emph{arXiv:2104.01233 [cs, eess]}, 2021.

\bibitem[Zanini et~al.(2018)Zanini, Congedo, Jutten, Said, and
  Berthoumieu]{zanini_transfer_2018}
Paolo Zanini, Marco Congedo, Christian Jutten, Salem Said, and Yannick
  Berthoumieu.
\newblock Transfer {Learning}: {A} {Riemannian} {Geometry} {Framework} {With}
  {Applications} to {Brain}–{Computer} {Interfaces}.
\newblock \emph{IEEE Transactions on Biomedical Engineering}, 65\penalty0
  (5):\penalty0 1107--1116, 2018.
\newblock \doi{10.1109/TBME.2017.2742541}.

\bibitem[Rodrigues et~al.(2019)Rodrigues, Jutten, and
  Congedo]{rodrigues_riemannian_2019}
Pedro Luiz~Coelho Rodrigues, Christian Jutten, and Marco Congedo.
\newblock Riemannian {Procrustes} {Analysis}: {Transfer} {Learning} for
  {Brain}–{Computer} {Interfaces}.
\newblock \emph{IEEE Transactions on Biomedical Engineering}, 66\penalty0
  (8):\penalty0 2390--2401, 2019.
\newblock \doi{10.1109/TBME.2018.2889705}.

\bibitem[Yair et~al.(2019)Yair, Ben-Chen, and Talmon]{yair_parallel_2019}
Or~Yair, Mirela Ben-Chen, and Ronen Talmon.
\newblock Parallel {Transport} on the {Cone} {Manifold} of {SPD} {Matrices} for
  {Domain} {Adaptation}.
\newblock \emph{IEEE Transactions on Signal Processing}, 67\penalty0
  (7):\penalty0 1797--1811, 2019.
\newblock \doi{10.1109/TSP.2019.2894801}.

\bibitem[Yong et~al.(2020)Yong, Huang, Meng, Hua, and
  Zhang]{yong_momentum_2020}
Hongwei Yong, Jianqiang Huang, Deyu Meng, Xiansheng Hua, and Lei Zhang.
\newblock Momentum {Batch} {Normalization} for {Deep} {Learning} with {Small}
  {Batch} {Size}.
\newblock In Andrea Vedaldi, Horst Bischof, Thomas Brox, and Jan-Michael Frahm,
  editors, \emph{Computer {Vision} – {ECCV} 2020}, volume 12357, pages
  224--240. Springer International Publishing, Cham, 2020.
\newblock ISBN 978-3-030-58609-6 978-3-030-58610-2.
\newblock \doi{10.1007/978-3-030-58610-2_14}.

\bibitem[He and Wu(2020)]{he_transfer_2020}
He~He and Dongrui Wu.
\newblock Transfer {Learning} for {Brain}–{Computer} {Interfaces}: {A}
  {Euclidean} {Space} {Data} {Alignment} {Approach}.
\newblock \emph{IEEE Transactions on Biomedical Engineering}, 67\penalty0
  (2):\penalty0 399--410, 2020.
\newblock \doi{10.1109/TBME.2019.2913914}.

\bibitem[Bakas et~al.(2022)Bakas, Ludwig, Barmpas, Bahri, Panagakis, Laskaris,
  Adamos, and Zafeiriou]{bakas_team_2022}
Stylianos Bakas, Siegfried Ludwig, Konstantinos Barmpas, Mehdi Bahri, Yannis
  Panagakis, Nikolaos Laskaris, Dimitrios~A. Adamos, and Stefanos Zafeiriou.
\newblock Team {Cogitat} at {NeurIPS} 2021: {Benchmarks} for {EEG} {Transfer}
  {Learning} {Competition}, 2022.

\bibitem[Bronstein et~al.(2017)Bronstein, Bruna, LeCun, Szlam, and
  Vandergheynst]{bronstein_geometric_2017}
Michael~M. Bronstein, Joan Bruna, Yann LeCun, Arthur Szlam, and Pierre
  Vandergheynst.
\newblock Geometric {Deep} {Learning}: {Going} beyond {Euclidean} data.
\newblock \emph{IEEE Signal Processing Magazine}, 34\penalty0 (4):\penalty0
  18--42, 2017.
\newblock \doi{10.1109/MSP.2017.2693418}.

\bibitem[Ju and Guan(2022)]{ju_deep_2022}
Ce~Ju and Cuntai Guan.
\newblock Deep {Optimal} {Transport} on {SPD} {Manifolds} for {Domain}
  {Adaptation}.
\newblock \emph{arXiv:2201.05745 [cs, eess]}, 2022.

\bibitem[Huang and Gool(2017)]{huang_riemannian_2017}
Zhiwu Huang and Luc~Van Gool.
\newblock A {Riemannian} {Network} for {SPD} {Matrix} {Learning}.
\newblock In \emph{Proceedings of the {Thirty}-{First} {AAAI} {Conference} on
  {Artificial} {Intelligence}}, {AAAI}'17, pages 2036--2042. AAAI Press, 2017.

\bibitem[Chang et~al.(2019)Chang, You, Seo, Kwak, and
  Han]{chang_domainspecific_2019}
Woong-Gi Chang, Tackgeun You, Seonguk Seo, Suha Kwak, and Bohyung Han.
\newblock Domain-{Specific} {Batch} {Normalization} for {Unsupervised} {Domain}
  {Adaptation}.
\newblock In \emph{Proceedings of the {IEEE}/{CVF} {Conference} on {Computer}
  {Vision} and {Pattern} {Recognition} ({CVPR})}, 2019.

\bibitem[Bhatia(2009)]{bhatia_positive_2009}
Rajendra Bhatia.
\newblock \emph{Positive definite matrices}.
\newblock Princeton university press, 2009.
\newblock ISBN 978-0-691-16825-8.

\bibitem[Absil et~al.(2009)Absil, Mahony, and
  Sepulchre]{absil_optimization_2009}
P-A Absil, Robert Mahony, and Rodolphe Sepulchre.
\newblock \emph{Optimization algorithms on matrix manifolds}.
\newblock Princeton University Press, 2009.

\bibitem[Pennec et~al.(2006)Pennec, Fillard, and
  Ayache]{pennec_riemannian_2006}
Xavier Pennec, Pierre Fillard, and Nicholas Ayache.
\newblock A {Riemannian} framework for tensor computing.
\newblock \emph{International Journal of computer vision}, 66\penalty0
  (1):\penalty0 41--66, 2006.

\bibitem[Karcher(1977)]{karcher_riemannian_1977}
Hermann Karcher.
\newblock Riemannian center of mass and mollifier smoothing.
\newblock \emph{Communications on pure and applied mathematics}, 30\penalty0
  (5):\penalty0 509--541, 1977.

\bibitem[Amari(2016)]{amari_information_2016}
Shun-ichi Amari.
\newblock \emph{Information geometry and its applications}, volume 194.
\newblock Springer, Tokyo, 1 edition, 2016.
\newblock ISBN 978-4-431-55978-8.

\bibitem[Brooks et~al.(2019)Brooks, Schwander, Barbaresco, Schneider, and
  Cord]{brooks_riemannian_2019}
Daniel Brooks, Olivier Schwander, Frederic Barbaresco, Jean-Yves Schneider, and
  Matthieu Cord.
\newblock Riemannian batch normalization for {SPD} neural networks.
\newblock In \emph{Advances in {Neural} {Information} {Processing} {Systems}},
  volume~32. Curran Associates, Inc., 2019.

\bibitem[Ioffe and Szegedy(2015)]{ioffe_batch_2015}
Sergey Ioffe and Christian Szegedy.
\newblock Batch normalization: {Accelerating} deep network training by reducing
  internal covariate shift.
\newblock In \emph{International conference on machine learning}, pages
  448--456. PMLR, 2015.

\bibitem[Kobler et~al.(2022)Kobler, Hirayama, and
  Kawanabe]{kobler_controlling_2022}
Reinmar~J. Kobler, Jun-ichiro Hirayama, and Motoaki Kawanabe.
\newblock Controlling {The} {Fréchet} {Variance} {Improves} {Batch}
  {Normalization} on the {Symmetric} {Positive} {Definite} {Manifold}.
\newblock In \emph{{ICASSP} 2022 - 2022 {IEEE} {International} {Conference} on
  {Acoustics}, {Speech} and {Signal} {Processing} ({ICASSP})}, pages
  3863--3867, Singapore, 2022. IEEE.
\newblock ISBN 978-1-66540-540-9.
\newblock \doi{10.1109/ICASSP43922.2022.9746629}.

\bibitem[Santurkar et~al.(2018)Santurkar, Tsipras, Ilyas, and
  Mądry]{santurkar_how_2018}
Shibani Santurkar, Dimitris Tsipras, Andrew Ilyas, and Aleksander Mądry.
\newblock How {Does} {Batch} {Normalization} {Help} {Optimization}?
\newblock In \emph{Proceedings of the 32nd {International} {Conference} on
  {Neural} {Information} {Processing} {Systems}}, {NIPS}'18, pages 2488--2498,
  Red Hook, NY, USA, 2018. Curran Associates Inc.

\bibitem[Ioffe(2017)]{ioffe_batch_2017}
Sergey Ioffe.
\newblock Batch {Renormalization}: {Towards} {Reducing} {Minibatch}
  {Dependence} in {Batch}-{Normalized} {Models}.
\newblock In \emph{Proceedings of the 31st {International} {Conference} on
  {Neural} {Information} {Processing} {Systems}}, {NIPS}'17, pages 1942--1950,
  Red Hook, NY, USA, 2017. Curran Associates Inc.
\newblock ISBN 978-1-5108-6096-4.

\bibitem[Ionescu et~al.(2015)Ionescu, Vantzos, and
  Sminchisescu]{ionescu_matrix_2015}
Catalin Ionescu, Orestis Vantzos, and Cristian Sminchisescu.
\newblock Matrix {Backpropagation} for {Deep} {Networks} with {Structured}
  {Layers}.
\newblock In \emph{2015 {IEEE} {International} {Conference} on {Computer}
  {Vision} ({ICCV})}, pages 2965--2973, Santiago, Chile, 2015. IEEE.
\newblock ISBN 978-1-4673-8391-2.
\newblock \doi{10.1109/ICCV.2015.339}.

\bibitem[Ho et~al.(2013)Ho, Cheng, Salehian, and Vemuri]{ho_recursive_2013}
Jeffrey Ho, Guang Cheng, Hesamoddin Salehian, and Baba Vemuri.
\newblock Recursive {Karcher} {Expectation} {Estimators} {And} {Geometric}
  {Law} of {Large} {Numbers}.
\newblock In Carlos~M. Carvalho and Pradeep Ravikumar, editors,
  \emph{Proceedings of the {Sixteenth} {International} {Conference} on
  {Artificial} {Intelligence} and {Statistics}}, volume~31 of \emph{Proceedings
  of {Machine} {Learning} {Research}}, pages 325--332, Scottsdale, Arizona,
  USA, 2013. PMLR.

\bibitem[Arsigny et~al.(2007)Arsigny, Fillard, Pennec, and
  Ayache]{arsigny_geometric_2007}
Vincent Arsigny, Pierre Fillard, Xavier Pennec, and Nicholas Ayache.
\newblock Geometric {Means} in a {Novel} {Vector} {Space} {Structure} on
  {Symmetric} {Positive}‐{Definite} {Matrices}.
\newblock \emph{SIAM Journal on Matrix Analysis and Applications}, 29\penalty0
  (1):\penalty0 328--347, 2007.
\newblock \doi{10.1137/050637996}.

\bibitem[Nunez and Srinivasan(2006)]{nunez_electric_2006}
Paul~L. Nunez and Ramesh Srinivasan.
\newblock \emph{Electric {Fields} of the {Brain}}.
\newblock Oxford University Press, 2006.
\newblock ISBN 978-0-19-505038-7.
\newblock \doi{10.1093/acprof:oso/9780195050387.001.0001}.

\bibitem[Michel et~al.(2004)Michel, Murray, Lantz, Gonzalez, Spinelli, and
  Grave De~Peralta]{Michel2004}
Christoph~M. Michel, Micah~M. Murray, Göran Lantz, Sara Gonzalez, Laurent
  Spinelli, and Rolando Grave De~Peralta.
\newblock {EEG} source imaging.
\newblock \emph{Clinical Neurophysiology}, 115\penalty0 (10), 2004.
\newblock \doi{10.1016/j.clinph.2004.06.001}.

\bibitem[Hyvärinen and Oja(2000)]{hyvarinen_independent_2000}
A.~Hyvärinen and E.~Oja.
\newblock Independent component analysis: algorithms and applications.
\newblock \emph{Neural Networks}, 13\penalty0 (4-5):\penalty0 411--430, 2000.
\newblock \doi{10.1016/S0893-6080(00)00026-5}.

\bibitem[Haufe et~al.(2014)Haufe, Meinecke, Görgen, Dähne, Haynes, Blankertz,
  and Bießmann]{haufe_on_2014}
Stefan Haufe, Frank Meinecke, Kai Görgen, Sven Dähne, John~Dylan Haynes,
  Benjamin Blankertz, and Felix Bießmann.
\newblock On the interpretation of weight vectors of linear models in
  multivariate neuroimaging.
\newblock \emph{NeuroImage}, 87:\penalty0 96--110, 2014.
\newblock \doi{10.1016/j.neuroimage.2013.10.067}.

\bibitem[Schirrmeister et~al.(2017)Schirrmeister, Springenberg, Fiederer,
  Glasstetter, Eggensperger, Tangermann, Hutter, Burgard, and
  Ball]{schirrmeister_deep_2017}
Robin~Tibor Schirrmeister, Jost~Tobias Springenberg, Lukas Dominique~Josef
  Fiederer, Martin Glasstetter, Katharina Eggensperger, Michael Tangermann,
  Frank Hutter, Wolfram Burgard, and Tonio Ball.
\newblock Deep learning with convolutional neural networks for {EEG} decoding
  and visualization: {Convolutional} {Neural} {Networks} in {EEG} {Analysis}.
\newblock \emph{Human Brain Mapping}, 38\penalty0 (11):\penalty0 5391--5420,
  2017.
\newblock \doi{10.1002/hbm.23730}.

\bibitem[Acharya et~al.(2018)Acharya, Huang, Pani~Paudel, and
  Van~Gool]{acharya_covariance_2018}
Dinesh Acharya, Zhiwu Huang, Danda Pani~Paudel, and Luc Van~Gool.
\newblock Covariance {Pooling} for {Facial} {Expression} {Recognition}.
\newblock In \emph{Proceedings of the {IEEE} {Conference} on {Computer}
  {Vision} and {Pattern} {Recognition} ({CVPR}) {Workshops}}, 2018.

\bibitem[Becigneul and Ganea(2019)]{becigneul_riemannian_2019}
Gary Becigneul and Octavian-Eugen Ganea.
\newblock Riemannian {Adaptive} {Optimization} {Methods}.
\newblock In \emph{International {Conference} on {Learning} {Representations}},
  2019.

\bibitem[Pfurtscheller and Neuper(2001)]{pfurtscheller_motor_2001}
G.~Pfurtscheller and C.~Neuper.
\newblock Motor imagery and direct brain-computer communication.
\newblock \emph{Proceedings of the IEEE}, 89\penalty0 (7):\penalty0 1123--1134,
  2001.
\newblock \doi{10.1109/5.939829}.

\bibitem[Dehais et~al.(2020)Dehais, Lafont, Roy, and
  Fairclough]{dehais_neuroergonomics_2020}
Frédéric Dehais, Alex Lafont, Raphaëlle Roy, and Stephen Fairclough.
\newblock A {Neuroergonomics} {Approach} to {Mental} {Workload}, {Engagement}
  and {Human} {Performance}.
\newblock \emph{Frontiers in Neuroscience}, 14:\penalty0 268, 2020.
\newblock \doi{10.3389/fnins.2020.00268}.

\bibitem[Donati et~al.(2016)Donati, Shokur, Morya, Campos, Moioli, Gitti,
  Augusto, Tripodi, Pires, Pereira, Brasil, Gallo, Lin, Takigami, Aratanha,
  Joshi, Bleuler, Cheng, Rudolph, and Nicolelis]{donati_long-term_2016}
Ana R.~C. Donati, Solaiman Shokur, Edgard Morya, Debora S.~F. Campos, Renan~C.
  Moioli, Claudia~M. Gitti, Patricia~B. Augusto, Sandra Tripodi, Cristhiane~G.
  Pires, Gislaine~A. Pereira, Fabricio~L. Brasil, Simone Gallo, Anthony~A. Lin,
  Angelo~K. Takigami, Maria~A. Aratanha, Sanjay Joshi, Hannes Bleuler, Gordon
  Cheng, Alan Rudolph, and Miguel A.~L. Nicolelis.
\newblock Long-{Term} {Training} with a {Brain}-{Machine} {Interface}-{Based}
  {Gait} {Protocol} {Induces} {Partial} {Neurological} {Recovery} in
  {Paraplegic} {Patients}.
\newblock \emph{Scientific Reports}, 6\penalty0 (1):\penalty0 30383, 2016.
\newblock \doi{10.1038/srep30383}.

\bibitem[Novak et~al.(2018)Novak, Sigrist, Gerig, Wyss, Bauer, Götz, and
  Riener]{novak_benchmarking_2018}
Domen Novak, Roland Sigrist, Nicolas~J. Gerig, Dario Wyss, René Bauer, Ulrich
  Götz, and Robert Riener.
\newblock Benchmarking {Brain}-{Computer} {Interfaces} {Outside} the
  {Laboratory}: {The} {Cybathlon} 2016.
\newblock \emph{Frontiers in Neuroscience}, 11:\penalty0 756, 2018.
\newblock \doi{10.3389/fnins.2017.00756}.

\bibitem[Tangermann et~al.(2012)Tangermann, Müller, Aertsen, Birbaumer, Braun,
  Brunner, Leeb, Mehring, Miller, Müller-Putz, Nolte, Pfurtscheller, Preissl,
  Schalk, Schlögl, Vidaurre, Waldert, and Blankertz]{Tangermann2012}
Michael Tangermann, Klaus-Robert Müller, Ad~Aertsen, Niels Birbaumer,
  Christoph Braun, Clemens Brunner, Robert Leeb, Carsten Mehring, Kai~J Miller,
  Gernot Müller-Putz, Guido Nolte, Gert Pfurtscheller, Hubert Preissl, Gerwin
  Schalk, Alois Schlögl, Carmen Vidaurre, Stephan Waldert, and Benjamin
  Blankertz.
\newblock Review of the {BCI} {Competition} {IV}.
\newblock \emph{Frontiers in Neuroscience}, 6, 2012.

\bibitem[Faller et~al.(2012)Faller, Vidaurre, Solis-Escalante, Neuper, and
  Scherer]{Faller2012}
Josef Faller, Carmen Vidaurre, Teodoro Solis-Escalante, Christa Neuper, and
  Reinhold Scherer.
\newblock Autocalibration and recurrent adaptation: towards a plug and play
  online {ERD}-{BCI}.
\newblock \emph{Neural Systems and Rehabilitation Engineering, IEEE
  Transactions on}, 20\penalty0 (3):\penalty0 313--319, 2012.

\bibitem[Lee et~al.(2019)Lee, Kwon, Kim, Kim, Lee, Williamson, Fazli, and
  Lee]{lee_eeg_2019}
Min-Ho Lee, O-Yeon Kwon, Yong-Jeong Kim, Hong-Kyung Kim, Young-Eun Lee, John
  Williamson, Siamac Fazli, and Seong-Whan Lee.
\newblock {EEG} dataset and {OpenBMI} toolbox for three {BCI} paradigms: an
  investigation into {BCI} illiteracy.
\newblock \emph{GigaScience}, 8\penalty0 (5):\penalty0 giz002, 2019.
\newblock \doi{10.1093/gigascience/giz002}.

\bibitem[Lehner et~al.(2020)Lehner, Robinson, Chouhan, Mihelj, Kratka,
  Debraine, Guan, and Wenderoth]{lehner_design_2020}
Rea Lehner, Neethu Robinson, Tushar Chouhan, Mihelj Mihelj, Ernest, Kratka
  Kratka, Paulina, Frédéric Debraine, Cuntai Guan, and Nicole Wenderoth.
\newblock Design considerations for long term non-invasive {Brain} {Computer}
  {Interface} training with tetraplegic {CYBATHLON} pilot: {CYBATHLON} 2020
  {Brain}-{Computer} {Interface} {Race} {Calibration} {Paradigms}.
\newblock 2020.
\newblock \doi{10.3929/ETHZ-B-000458693}.
\newblock URL \url{http://hdl.handle.net/20.500.11850/458693}.

\bibitem[Stieger et~al.(2021)Stieger, Engel, and He]{stieger_continuous_2021}
James~R. Stieger, Stephen~A. Engel, and Bin He.
\newblock Continuous sensorimotor rhythm based brain computer interface
  learning in a large population.
\newblock \emph{Scientific Data}, 8\penalty0 (1):\penalty0 98, 2021.
\newblock \doi{10.1038/s41597-021-00883-1}.

\bibitem[Hehenberger et~al.(2021)Hehenberger, Kobler, Lopes-Dias, Srisrisawang,
  Tumfart, Uroko, Torke, and Müller-Putz]{hehenberger_long-term_2021}
Lea Hehenberger, Reinmar~J. Kobler, Catarina Lopes-Dias, Nitikorn Srisrisawang,
  Peter Tumfart, John~B. Uroko, Paul~R. Torke, and Gernot~R. Müller-Putz.
\newblock Long-term mutual training for the {CYBATHLON} {BCI} {Race} with a
  tetraplegic pilot: a case study on inter-session transfer and intra-session
  adaptation.
\newblock \emph{Frontiers in Human Neuroscience}, 2021.
\newblock \doi{10.3389/fnhum.2021.635777}.

\bibitem[Hinss et~al.(2021)Hinss, Darmet, Somon, Jahanpour, Lotte, Ladouce, and
  Roy]{hinss_eegdata_2021}
Marcel~F. Hinss, Ludovic Darmet, Bertille Somon, Emilie Jahanpour, Fabien
  Lotte, Simon Ladouce, and Raphaëlle~N. Roy.
\newblock An {EEG} dataset for cross-session mental workload estimation:
  {Passive} {BCI} competition of the {Neuroergonomics} {Conference} 2021.
\newblock 2021.
\newblock \doi{10.5281/ZENODO.5055046}.

\bibitem[Gramfort(2013)]{gramfort_meg_2013}
Alexandre Gramfort.
\newblock {MEG} and {EEG} data analysis with {MNE}-{Python}.
\newblock \emph{Frontiers in Neuroscience}, 7, 2013.
\newblock \doi{10.3389/fnins.2013.00267}.

\bibitem[{Kai Keng Ang} et~al.(2008){Kai Keng Ang}, {Zhang Yang Chin}, {Haihong
  Zhang}, and {Cuntai Guan}]{kai_keng_ang_filter_2008}
{Kai Keng Ang}, {Zhang Yang Chin}, {Haihong Zhang}, and {Cuntai Guan}.
\newblock Filter {Bank} {Common} {Spatial} {Pattern} ({FBCSP}) in
  {Brain}-{Computer} {Interface}.
\newblock In \emph{2008 {IEEE} {International} {Joint} {Conference} on {Neural}
  {Networks} ({IEEE} {World} {Congress} on {Computational} {Intelligence})},
  pages 2390--2397, Hong Kong, China, 2008. IEEE.
\newblock ISBN 978-1-4244-1820-6.
\newblock \doi{10.1109/IJCNN.2008.4634130}.

\bibitem[Yair et~al.(2020)Yair, Dietrich, Talmon, and
  Kevrekidis]{yair_domain_2020}
Or~Yair, Felix Dietrich, Ronen Talmon, and Ioannis~G. Kevrekidis.
\newblock Domain {Adaptation} with {Optimal} {Transport} on the {Manifold} of
  {SPD} matrices.
\newblock \emph{arXiv:1906.00616 [cs, stat]}, 2020.

\bibitem[Lawhern et~al.(2018)Lawhern, Solon, Waytowich, Gordon, Hung, and
  Lance]{lawhern_eegnet_2018}
Vernon~J Lawhern, Amelia~J Solon, Nicholas~R Waytowich, Stephen~M Gordon,
  Chou~P Hung, and Brent~J Lance.
\newblock {EEGNet}: a compact convolutional neural network for {EEG}-based
  brain–computer interfaces.
\newblock \emph{Journal of Neural Engineering}, 15\penalty0 (5):\penalty0
  056013, 2018.
\newblock \doi{10.1088/1741-2552/aace8c}.

\bibitem[Paszke et~al.(2019)Paszke, Gross, Massa, Lerer, Bradbury, Chanan,
  Killeen, Lin, Gimelshein, Antiga, Desmaison, Kopf, Yang, DeVito, Raison,
  Tejani, Chilamkurthy, Steiner, Fang, Bai, and Chintala]{paszke_pytorch_2019}
Adam Paszke, Sam Gross, Francisco Massa, Adam Lerer, James Bradbury, Gregory
  Chanan, Trevor Killeen, Zeming Lin, Natalia Gimelshein, Luca Antiga, Alban
  Desmaison, Andreas Kopf, Edward Yang, Zachary DeVito, Martin Raison, Alykhan
  Tejani, Sasank Chilamkurthy, Benoit Steiner, Lu~Fang, Junjie Bai, and Soumith
  Chintala.
\newblock {PyTorch}: {An} {Imperative} {Style}, {High}-{Performance} {Deep}
  {Learning} {Library}.
\newblock In H.~Wallach, H.~Larochelle, A.~Beygelzimer, F.~d' Alché-Buc,
  E.~Fox, and R.~Garnett, editors, \emph{Advances in {Neural} {Information}
  {Processing} {Systems} 32}, pages 8024--8035. Curran Associates, Inc., 2019.

\bibitem[Pedregosa et~al.(2011)Pedregosa, Varoquaux, Gramfort, Michel, Thirion,
  Grisel, Blondel, Prettenhofer, Weiss, Dubourg, Vanderplas, Passos,
  Cournapeau, Brucher, Perrot, and Duchesnay]{pedregosa_scikitlearn_2011}
F.~Pedregosa, G.~Varoquaux, A.~Gramfort, V.~Michel, B.~Thirion, O.~Grisel,
  M.~Blondel, P.~Prettenhofer, R.~Weiss, V.~Dubourg, J.~Vanderplas, A.~Passos,
  D.~Cournapeau, M.~Brucher, M.~Perrot, and E.~Duchesnay.
\newblock Scikit-learn: {Machine} {Learning} in {Python}.
\newblock \emph{Journal of Machine Learning Research}, 12:\penalty0 2825--2830,
  2011.

\bibitem[Tietz et~al.(2017)Tietz, Fan, Nouri, Bossan, and {skorch
  Developers}]{tietz_skorch_2017}
Marian Tietz, Thomas~J. Fan, Daniel Nouri, Benjamin Bossan, and {skorch
  Developers}.
\newblock \emph{skorch: {A} scikit-learn compatible neural network library that
  wraps {PyTorch}}.
\newblock 2017.
\newblock URL \url{https://skorch.readthedocs.io/en/stable/}.

\bibitem[Kochurov et~al.(2020)Kochurov, Karimov, and
  Kozlukov]{kochurov_geoopt_2020}
Max Kochurov, Rasul Karimov, and Serge Kozlukov.
\newblock Geoopt: {Riemannian} {Optimization} in {PyTorch}, 2020.
\newblock \_eprint: 2005.02819.

\bibitem[Alexandre(2022)]{alexandre_pyriemann_2022}
Barachant Alexandre.
\newblock {pyRiemann}, 2022.
\newblock URL \url{https://github.com/pyRiemann/pyRiemann}.

\bibitem[Townsend et~al.(2016)Townsend, Koep, and
  Weichwald]{townsend_pymanopt_2016}
James Townsend, Niklas Koep, and Sebastian Weichwald.
\newblock Pymanopt: A python toolbox for optimization on manifolds using
  automatic differentiation.
\newblock \emph{Journal of Machine Learning Research}, 17\penalty0
  (137):\penalty0 1--5, 2016.

\bibitem[Campbell(2009)]{campbell_eeg_2009}
Ian~G. Campbell.
\newblock {EEG} {Recording} and {Analysis} for {Sleep} {Research}.
\newblock \emph{Current Protocols in Neuroscience}, 49\penalty0 (1):\penalty0
  10.2.1--10.2.19, 2009.
\newblock \doi{10.1002/0471142301.ns1002s49}.

\bibitem[Banville et~al.(2020)Banville, Chehab, Hyvärinen, Engemann, and
  Gramfort]{banville_uncovering_2020}
Hubert Banville, Omar Chehab, Aapo Hyvärinen, Denis-Alexander Engemann, and
  Alexandre Gramfort.
\newblock Uncovering the structure of clinical {EEG} signals with
  self-supervised learning.
\newblock \emph{Journal of Neural Engineering}, 2020.
\newblock \doi{10.1088/1741-2552/abca18}.

\bibitem[Acharya et~al.(2013)Acharya, Sree, Swapna, Martis, and
  Suri]{acharya_automated_2013}
U.~Rajendra Acharya, S.~Vinitha Sree, G.~Swapna, Roshan~Joy Martis, and
  Jasjit~S. Suri.
\newblock Automated {EEG} analysis of epilepsy: {A} review.
\newblock \emph{Knowledge-Based Systems}, 45:\penalty0 147--165, 2013.
\newblock \doi{https://doi.org/10.1016/j.knosys.2013.02.014}.

\bibitem[Nuwer et~al.(2005)Nuwer, Hovda, Schrader, and
  Vespa]{nuwer_routine_2005}
Marc~R. Nuwer, David~A. Hovda, Lara~M. Schrader, and Paul~M. Vespa.
\newblock Routine and quantitative eeg in mild traumatic brain injury.
\newblock \emph{Clinical Neurophysiology}, 116\penalty0 (9):\penalty0
  2001--2025, 2005.
\newblock \doi{10.1016/j.clinph.2005.05.008}.

\bibitem[Gemein et~al.(2020)Gemein, Schirrmeister, Chrabaszcz, Wilson,
  Boedecker, Schulze-Bonhage, Hutter, and
  Ball]{gemein_machine-learning-based_2020}
Lukas~A.W. Gemein, Robin~T. Schirrmeister, Patryk Chrabaszcz, Daniel Wilson,
  Joschka Boedecker, Andreas Schulze-Bonhage, Frank Hutter, and Tonio Ball.
\newblock Machine-learning-based diagnostics of {EEG} pathology.
\newblock \emph{NeuroImage}, 220:\penalty0 117021, 2020.
\newblock \doi{10.1016/j.neuroimage.2020.117021}.

\bibitem[Ganin et~al.(2016)Ganin, Ustinova, Ajakan, Germain, Larochelle,
  Laviolette, March, and Lempitsky]{ganin_domain-adversarial_2016}
Yaroslav Ganin, Evgeniya Ustinova, Hana Ajakan, Pascal Germain, Hugo
  Larochelle, François Laviolette, Mario March, and Victor Lempitsky.
\newblock Domain-{Adversarial} {Training} of {Neural} {Networks}.
\newblock \emph{Journal of Machine Learning Research}, 17\penalty0
  (59):\penalty0 1--35, 2016.

\end{thebibliography}







\section*{Checklist}


\begin{enumerate}

\item For all authors...
\begin{enumerate}
  \item Do the main claims made in the abstract and introduction accurately reflect the paper's contributions and scope?
    \answerYes{}
  \item Did you describe the limitations of your work?
    \answerYes{See section \ref{sec:discussion}}
  \item Did you discuss any potential negative societal impacts of your work?
    \answerYes{See section \ref{sec:discussion}}
  \item Have you read the ethics review guidelines and ensured that your paper conforms to them?
    \answerYes{}
\end{enumerate}

\item If you are including theoretical results...
\begin{enumerate}
  \item Did you state the full set of assumptions of all theoretical results?
    \answerYes{See paragraph 4 in section\,\ref{par:spdmbn_converges_to_frechetmean}.}
        \item Did you include complete proofs of all theoretical results?
    \answerYes{See appendix\,\ref{appendix:proof} in the supplementary material.}
\end{enumerate}

\item If you ran experiments...
\begin{enumerate}
  \item Did you include the code, data, and instructions needed to reproduce the main experimental results (either in the supplemental material or as a URL)?
    \answerYes{See Supplementary Material.}
  \item Did you specify all the training details (e.g., data splits, hyperparameters, how they were chosen)?
    \answerYes{See paragraph 2 in section \ref{par:experiments:evaluation} for a brief summary and appendix \ref{appendix:methods} for details.}
        \item Did you report error bars (e.g., with respect to the random seed after running experiments multiple times)?
    \answerYes{We provide error bars with respect to the variability across sessions and subjects.}
        \item Did you include the total amount of compute and the type of resources used (e.g., type of GPUs, internal cluster, or cloud provider)?
    \answerYes{See paragraph 3 in section \ref{par:experiments:softhardware}.}
\end{enumerate}

\item If you are using existing assets (e.g., code, data, models) or curating/releasing new assets...
\begin{enumerate}
  \item If your work uses existing assets, did you cite the creators?
    \answerYes{See paragraph 1 in section \ref{par:experiments:datasets}.}
  \item Did you mention the license of the assets?
    \answerYes{See appendix \ref{appendix:methods:datasets} in the Supplementary material.}
  \item Did you include any new assets either in the supplemental material or as a URL?
    \answerNo{}
  \item Did you discuss whether and how consent was obtained from people whose data you're using/curating?
    \answerYes{See paragraph 1 in section \ref{par:experiments:datasets}.}
  \item Did you discuss whether the data you are using/curating contains personally identifiable information or offensive content?
    \answerYes{See paragraph 1 in section \ref{par:experiments:datasets}.}
\end{enumerate}

\item If you used crowdsourcing or conducted research with human subjects...
\begin{enumerate}
  \item Did you include the full text of instructions given to participants and screenshots, if applicable?
    \answerNA{}
  \item Did you describe any potential participant risks, with links to Institutional Review Board (IRB) approvals, if applicable?
    \answerNA{}
  \item Did you include the estimated hourly wage paid to participants and the total amount spent on participant compensation?
    \answerNA{}
\end{enumerate}

\end{enumerate}


\appendix

\section{Supplementary Material}

\subsection{Proof of proposition \ref{prop:spdbnbound}}
\label{appendix:proof}

In the proof, we will use Theorem 2 of \cite{ho_recursive_2013} which relates the distance between interpolated points along the geodesic $\vec{R} \#_\gamma \vec{S}$, connecting points $\vec{R}$ and $\vec{S}$, and point $\vec{T}$ to the distances between $\vec{R}$, $\vec{S}$ and $\vec{T}$.
Formally, for all $\vec{R},\vec{S}, \vec{T} \in \mathcal{S}_D^+$ we have

\begin{equation}
  \delta_{}^2(\vec{R} \#_\gamma \vec{S}, \vec{T}) \le (1-\gamma) \delta_{}^2(\vec{R}, \vec{T}) + \gamma \delta_{}^2(\vec{S}, \vec{T}) - \gamma (1-\gamma) \delta_{}^2(\vec{R}, \vec{S})
  \label{eq:ho2013:thm3}
\end{equation}

As a last ingredient for the proof, we use Proposition 1 of \cite{ho_recursive_2013}, which states that for a random variable $\vec{U}$, following distribution $P_{\mathcal{U}}$ defined on $\mathcal{S}_D^+$ with Fréchet mean $\vec{G}_{\mathcal{U}}$, we have for any point $\vec{V} \in \mathcal{S}_D^+$:

\begin{equation}
  \mathbb{E} \{ \delta_{}^2(\vec{U}, \vec{V}) \} \ge \underbrace{\int \delta_{}^2(\vec{u}, \vec{G}_{\mathcal{U}}) dP_{\mathcal{U}}(\vec{u})}_{=: \mrm{Var}(\vec{U}) } + \delta_{}^2(\vec{V}, \vec{G}_{\mathcal{U}})
  \label{eq:ho2013:prop1}
\end{equation}

which means that the expected distance between $\vec{V}$ and $\vec{U}$ is bounded from below by the variance of $\vec{U}$ and the distance between $\vec{V}$ and its Fréchet mean $\vec{G}_{\mathcal{U}}$.

Let us quickly repeat the definitions, assumptions and proposition \ref{prop:spdbnbound}.
We define a dataset containing the latent representations generated with feature set $\theta_k$ as $\mathcal{Z}_{\theta_k} = \{ f_{\theta_k} (\vec{x}) | \vec{x} \in \mathcal{T} \}$, and a minibatch of $M$ iid samples drawn from $\mathcal{Z}_{\theta_k}$ at training step $k$ as $\mathcal{B}_k$.
We denote the Fréchet mean of $\mathcal{Z}_{\theta_k}$ as $\vec{G}_{\theta_k}$, the estimated Fréchet mean, defined in (\ref{eq:spdrunningmean}), as $\vec{G}_k$, and the estimated batch mean as $\vec{B}_k$. Since the batches are drawn randomly, we consider $\vec{B}_k$ and $\vec{G}_k$ as random variables.

We assume that the variance $\mrm{Var}_{\theta_{k}}(\vec{B}_{k-1}) = \mathbb{E}_{\vec{B}_{k-1}} \{ \delta^2(\vec{B}_{k-1}, \vec{G}_{\theta_{k}}) \}$ of the previous batch mean $\vec{B}_{k-1}$ with respect to the current Fréchet mean $\vec{G}_{\theta_k}$ is bounded by the current variance $\mrm{Var}_{\theta_{k}}(\vec{B}_{k})$ and the norm of the difference in the parameters

\begin{equation}
  \mrm{Var}_{\theta_k}(\vec{B}_{k-1}) \le (1 + || \theta_k - \theta_{k-1} ||) \mrm{Var}_{\theta_k}(\vec{B}_{k}) 
  \label{eq:proof:assumption1}
\end{equation}

Proposition \ref{prop:spdbnbound} states then that the variance of the running mean $\mrm{Var}_{\theta_k}(\vec{G}_k)$ is bounded by  
\begin{equation}
  \mrm{Var}_{\theta_k}(\vec{G}_k) \le \mrm{Var}_{\theta_k}(\vec{B}_k)
\end{equation}
over training steps $k$, if
\begin{equation}
  || \theta_k - \theta_{k-1} || \le  \frac{1- \gamma^2}{(1-\gamma)^2} - 1
\end{equation}
holds true.

\begin{proof}[Proof of Proposition 1]
  We prove Proposition \ref{prop:spdbnbound} via induction.
  We assume that the variance $\mrm{Var}_{\theta_{k}}(\vec{G}_{k-1})$, that is the expected distance between the running mean $\vec{G}_{k-1}$ and the Fréchet mean $\vec{G}_{\theta_{k}}$, is bounded by the variance of the batch mean $\vec{B}_{k-1}$:
  
  \begin{equation}
    \mrm{Var}_{\theta_{k}}(\vec{G}_{k-1}) = \mathbb{E}_{\vec{G}_{k-1}} \{ \delta_{}^2(\vec{G}_{k-1}, \vec{G}_{\theta_{k}}) \} \le \mrm{Var}_{\theta_{k}}(\vec{B}_{k-1})
    \label{eq:proof:assumption}
  \end{equation}

  and show that this also holds for $\vec{G}_{k}$ and $\vec{B}_{k}$.
  The assumption is trivially satisfied for $\vec{G}_{0} = \vec{B}_{0}$.
  We start the induction step with (\ref{eq:ho2013:thm3}) and apply it to the update rule for the running estimate $\vec{G}_k$.
  As a result, we have

  \begin{equation}
    \delta^2(\vec{G}_k, \vec{G}_{\theta_k}) \le (1-\gamma) \delta^2(\vec{G}_{k-1}, \vec{G}_{\theta_k}) + \gamma \delta^2(\vec{B}_k, \vec{G}_{\theta_k}) - \gamma (1-\gamma) \delta^2(\vec{G}_{k-1}, \vec{B}_k)
    \label{eq:updateinequality}
  \end{equation}

  where we used $\vec{G}_k = \vec{G}_{k-1} \#_\gamma \vec{B}_k$, as defined in algorithm \ref{algo:spdmbn}.
  Taking the expectations for the random variables $\vec{G}_k$, $\vec{G}_{k-1}$ and $\vec{B}_k$, we get

  \begin{equation}
    \mrm{Var}_{\theta_k}(\vec{G}_k) \le (1-\gamma) \mrm{Var}_{\theta_k}(\vec{G}_{k-1}) + \gamma \mrm{Var}_{\theta_k}(\vec{B}_{k}) - \gamma (1-\gamma) \mathbb{E}_{\vec{G}_{k-1}} \{  \mathbb{E}_{\vec{B}_{k}} \{  \delta^2(\vec{G}_{k-1}, \vec{B}_k) \} \}
  \end{equation}

  Using (\ref{eq:ho2013:prop1}) to simplify the last term, we obtain

  \begin{align}
    \mrm{Var}_{\theta_k}(\vec{G}_k) &\le (1-\gamma) \mrm{Var}_{\theta_k}(\vec{G}_{k-1}) + \gamma \mrm{Var}_{\theta_k}(\vec{B}_{k}) - \gamma (1-\gamma) \left( \mrm{Var}_{\theta_k}(\vec{G}_{k-1}) + \mrm{Var}_{\theta_k}(\vec{B}_{k}) \right) \\
    & \le (1-\gamma)^2 \mrm{Var}_{\theta_k}(\vec{G}_{k-1}) + \gamma^2 \mrm{Var}_{\theta_k}(\vec{B}_{k})
  \end{align}

  Applying assumptions (\ref{eq:proof:assumption}) and (\ref{eq:proof:assumption1}), we get

  \begin{align}
    \mrm{Var}_{\theta_k}(\vec{G}_k)  & \overset{(\ref{eq:proof:assumption})}{\le} (1-\gamma)^2  \mrm{Var}_{\theta_k}(\vec{B}_{k-1}) + \gamma^2 \mrm{Var}_{\theta_k}(\vec{B}_{k}) \\
    & \overset{(\ref{eq:proof:assumption1})}{\le} \left[  (1-\gamma)^2 (1 + || \theta_k - \theta_{k-1} ||) + \gamma^2 \right] \mrm{Var}_{\theta_k}(\vec{B}_{k})
  \end{align}

  The resulting inequality holds true, for
  \begin{equation}
    (1-\gamma)^2 (1 + || \theta_k - \theta_{k-1} ||) + \gamma^2 \overset{!}{\le} 1 \Leftrightarrow || \theta_k - \theta_{k-1} || \overset{!}{\le} \frac{1-\gamma^2}{(1-\gamma)^2} - 1
  \end{equation}
  and, in turn, results in feasible bounds for the parameter updates for fixed $\gamma \in (0, 1)$.
  This concludes the proof.
\end{proof}

\subsection{Supplementary Methods}
\label{appendix:methods}

\subsubsection{Datasets}
\label{appendix:methods:datasets}

A summary of the datasets' key attributes is listed in Table\,\ref{tab:datasets}.
The datasets contain a diverse sample of 154 human subjects, whose data was acquired in Europe (38 subjects; BNCI2014001,BNCI2015001,Lehner2021,Hehenberger2021,Hinss2021), Asia (54 subjects; Lee2019) and North America (62 subjects; Stieger2021).

\begin{table}[p]
  \caption{Dataset attributes. The epoch indices the temporal window after a task cue (time relative to cue onset) that was extracted from continuous EEG data.}
  \label{tab:datasets}
  \centering
  \footnotesize
  \begin{tabular}{lrrrrrr}
    \toprule
    {} &         epoch  &  sampling &           channels &  subjects &  sessions & obsevations  \\
    dataset         &      (s)     &  rate (Hz)&            \#        &     \#         &     \#         & \# (per session)  \\
    \midrule
    BNCI2014001       &  0.5 - 3.5 &       250 &                 22 &          9 &              2 & 288\\
    BNCI2015001       &  1.0 - 4.0 &       256 &                 13 &         12 &            2-3 & 200\\
    Lee2019        &  1.0 - 3.5 &       250 &  20\tablefootnote{We used 20 channels that covered sensorimotor areas.} &        54 &  2 &  100 \\
    Stieger2021    &  1.0 - 3.0 &       250 &  34\tablefootnote{We used 34 channels that covered sensorimotor areas.} &        62 &  4-8\tablefootnote{To reduce computation time, only data from the 4th to last session were considered (inter-session) or last session (inter-subject).} & 390 \\
    Lehner2021   &  0.5 - 2.5 &       250 &                 60 &          1 &              7 & 61\\
    Hehenberger2021  &  1.0 - 3.0 &       250 &                 32 &          1 &              26 & 105\\
    Hinss2021      &  0.0 - 2.0 &       250 &      30\tablefootnote{We used 30 channels with a dense coverage in frontal areas.} &        15 &  2 & 447 \\
    \midrule
    \end{tabular}
    \begin{tabular}{lrrrr}
    {} &  classes &    license &     identifier  &  linked    \\
    dataset     \#    &   &      &  & publication\\
    \midrule
    BNCI2014001  & 4  & \href{http://creativecommons.org/licenses/by-nd/4.0/}{CC BY-ND 4.0}    &  \href{http://bnci-horizon-2020.eu/database/data-sets}{001-2014} & \cite{Tangermann2012}   \\
    BNCI2015001  & 2  & \href{http://creativecommons.org/licenses/by-nc-nd/4.0/}{CC BY-NC-ND 4.0} &  \href{http://bnci-horizon-2020.eu/database/data-sets}{001-2015} & \cite{Faller2012}  \\
    Lee2019   &   2   & unspecified     &  \href{https://gigadb.org/dataset/100542}{100542} & \cite{lee_eeg_2019}   \\
    Stieger2021 &  4  & \href{https://creativecommons.org/licenses/by/4.0/}{CC BY 4.0}       &  \href{https://doi.org/10.6084/m9.figshare.13123148.v1}{m9.figshare.13123148.v1} & \cite{stieger_continuous_2021}   \\
    Lehner2021   &  2    & \href{http://rightsstatements.org/page/InC-NC/1.0/}{InC-NC}         &  \href{https://doi.org/10.3929/ethz-b-000458693}{10.3929/ethz-b-000458693} & \cite{lehner_design_2020}   \\
    Hehenberger2021 &4 &individual\tablefootnote{The dataset is shared on an individual basis by the authors of \cite{hehenberger_long-term_2021}.}      &  - & \cite{hehenberger_long-term_2021}   \\
    Hinss2021  & 3    & \href{https://creativecommons.org/licenses/by-sa/4.0/}{CC BY-SA 4.0}    &  \href{https://doi.org/10.5281/zenodo.5055046}{10.5281/zenodo.5055046} & \cite{hinss_eegdata_2021}  \\
    \bottomrule
    \end{tabular}
\end{table}

\subsubsection{Preprocessing}
\label{appendix:methods:preprocessing}
Depending on the dataset, either all or a subset of EEG channels was selected, and then resampled along the temporal dimension to a sampling rate of either 250 or 256 Hz. (see Table \ref{tab:datasets}).
Thereafter, an infinite impulse response (IIR) bandpass filter was used to extract EEG activity in the 4 to 36 Hz range (4th order Butterworth filter, 4 and 36 Hz cut-off frequencies, zero-phase).
Some baseline methods required spectrally resolved input data.
For these, we applied a bank of 8 filters with similar parameters except for the cut-off frequencies ([4, 8], [8, 12], ..., [32, 36] Hz). 
Finally, short epochs (=segments) were extracted (see Table \ref{tab:datasets}) relative to the task cues (=labels).
The labeled data were then extended with a domain index (= unique integer associated to one session of one subject).

\subsubsection{Baseline methods}
\label{appendix:methods:baseline}
We considered several established and SoA baseline methods which were previously applied to inter-session/-subject TL.
They can be broadly categorized as component based, Riemannian geometry aware or deep learning which we denote component, geometric and end-to-end, respectively. 
For the component category, we considered the popular filter-bank common spatial patterns (FBCSP+SVM) \cite{kai_keng_ang_filter_2008} and a variant \cite{hehenberger_long-term_2021}, designed for MSMTUDA, that applies domain-specific standardization (DSS) to features before classification, denoted FBCSP+DSS+LDA.
The geometric category was represented by TSM+SVM \cite{Barachant2012}, a spectrally resolved variant \cite{kobler_interpretation_2021} denoted FB+TSM+LR (which was also used as domain-specific baseline method).
We additionally considered two MSUDA methods, denoted URPA+MDM and SPDOT+TSM+SVM here, that align SPD observations (=spatial covariance matrices) of different domains.
The former uses Riemannian Procrustes Analysis (RPA) \cite{rodrigues_riemannian_2019} to align domains, and the latter optimal transport (OT) on $\mathcal{S}^+_D$ \cite{yair_domain_2020}.
The end-to-end category was represented by EEGNet \cite{lawhern_eegnet_2018} and ShConvNet \cite{schirrmeister_deep_2017} two convolutional neural network architectures specifically designed for EEG data.
We additionally considered variants \cite{ozdenizci_learning_2020} that use domain-adversarial neural networks (DANN) \cite{ganin_domain-adversarial_2016} to learn domain-invariant latent representations.

\subsubsection{Domain-specific reference method}
\label{appendix:methods:dsreference}
Due to the success of TSM models \cite{jayaram_moabb_2018,lotte_review_2018}, we considered a spectrally resolved model \cite{sabbagh_manifold-regression_2019,kobler_interpretation_2021} which consisted of a filter-bank to separate activity of canonical frequency bands. 
For each frequency band, PCA was used to reduce the spatial dimensionality and TSM to project the SPD features to the Euclidean vector space. 
Finally, all features were pooled and submitted to a penalized logistic regression classifier. For further details, see \cite{kobler_interpretation_2021}.

\subsubsection{TSMNet}
\label{appendix:methods:tsmnet}

\begin{figure}[p]
  \centering
  \includegraphics[width=0.7\textwidth]{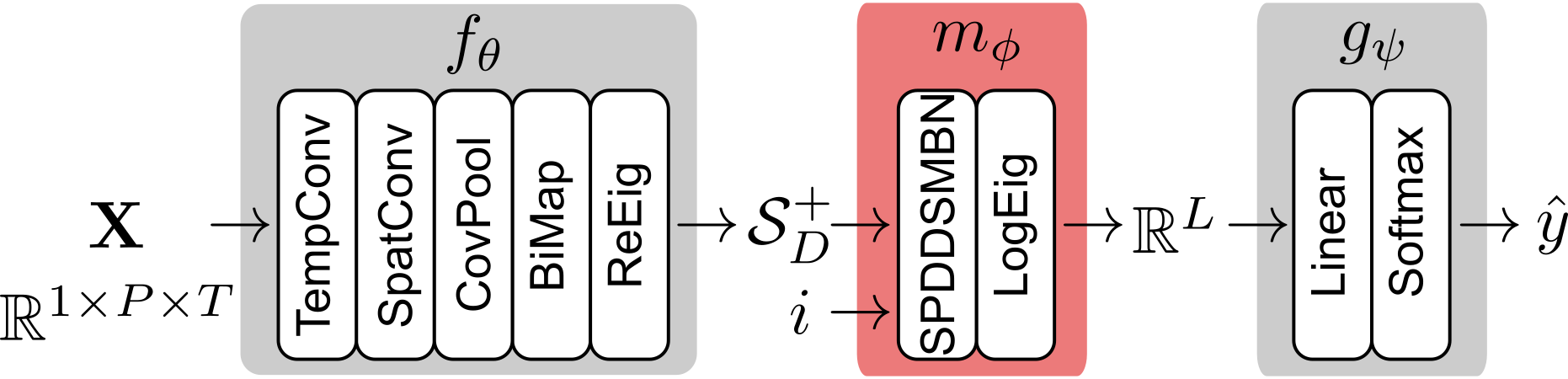}
  \caption{Overview of the TSMNet architecture.
  The network uses observations $\vec{X}$ and the associated domain index $i$ to estimate the target label $\hat{y}$.
  }\label{fig:architecture}
\end{figure}

\paragraph{Architecture}
The architecture of TSMNet is outlined in Figure \ref{fig:architecture} and detailed in Table \ref{tab:architecture}.
The feature extractor $f_\theta$ comprises two convolutional layers, followed by covariance pooling \cite{acharya_covariance_2018}, BiMap \cite{huang_riemannian_2017} and ReEig \cite{huang_riemannian_2017} layers.
The first convolutional layer performs convolution along the temporal dimension; implementing a finite impulse response (FIR) filter bank (4 filters) with learnable parameters.
The second convolutional layer applies spatio-spectral filters (40 filters) along the spatial and convolutional channel dimensions.
Covariance pooling is then applied along the temporal dimension.
A subsequent BiMap layer projects covariance matrices to a subspace via a bilinear mapping (i.e, $\mrm{BiMap}(\vec{Z}) = \vec{W}^T_\theta \vec{Z} \vec{W}_\theta $) where the parameter matrix $\vec{W}_\theta$ is constrained to have orthogonal rows (i.e., $\vec{W}_\theta \in \{ \vec{U} \in \mathbb{R}^{I \times O} : \vec{U}^T \vec{U} = \vec{I}_O, I \ge O \}$).
Next, a ReEig layer rectifies all eigenvalues, lower than a threshold $\epsilon = 10^{-4}$ (i.e, $\mrm{ReEig}(\vec{Z}) = \vec{U} \mathrm{max}(\vec{\Sigma},\epsilon \vec{I}) \vec{U}^T$ with $[\vec{\Sigma},\vec{U}] = \mathrm{eig}(\vec{Z})$).\\
Domain-specific tangent space mapping $m_\phi$ is implemented via combining SPDDSMBN and LogEig layers.
In order for $m_{\phi}$ to align the domain data and compute TSM, we use SPDDSMBN (\ref{eq:spddsmbn}) with shared parameters (i.e., $\vec{G}_{\phi_i} = \vec{G}_{\phi} = \vec{I}, \nu_{\phi_i} = \nu_{\phi}$) in (\ref{eq:tsmlayer}).
The classifier $g_{\psi}$ was parametrized as a linear layer with softmax activations. 

\paragraph{Parameter estimation}
We used the cross entropy loss as training objective, and the standard backpropagation framework with extensions for structured matrices \cite{ionescu_matrix_2015} and manifold-constrained gradients \cite{absil_optimization_2009} to propagate gradients through the layers of TSMNet.
Gradients were estimated with mini-batches of fixed size (50 observations; 10 per domain; 5 distinct domains) and converted into parameter updates with the Riemannian ADAM optimizer \cite{becigneul_riemannian_2019} ($10^{-3}$ learning rate, $10^{-4}$ weight decay applied to unconstrained parameters; $\beta_1 = 0.9, \beta_2 = 0.999$).

For every MSMTUDA problem, comprising source and target domain sets, we split the source domains' data into training and validation sets (80\%/20\% splits; randomized; stratified across domains and labels) and repeatedly iterated through the training set for 50 epochs via exhaustive minibatch sampling.
As required by SPDMBN, we implemented a decaying schedule (over epochs) for the training momentum parameter $\gamma_{train}(k)$, defined in (\ref{eq:momentumschedule}), with $\gamma_{train}(0) = 1$ and $\gamma_{min} = 0.2$ attained at epoch $K = 40$.
During training, we monitored the loss on the validation data (at the end of every epoch).
Post training, the model with minimal loss on the validation data was selected.
For each target domain, the associated data was then passed through this model to estimate the labels.
During the forward pass, the domain's normalization statistics within the SPDMBN layers were computed by solving (\ref{eq:geommean}) with the Karcher flow algorithm \cite{karcher_riemannian_1977}.

\begin{table}[p]
  \centering
  \caption{TSMNet architecture details.
  The letters P, T and C refer to the number of input channels, temporal samples and classes.
  }\label{tab:architecture}
  \footnotesize
  \begin{tabular}{lrrrrll}
    \toprule
    Block & Input (dim)    & Output (dim)     &    Parameter (dim) &           Operation &                      Note \\
    \midrule
    TempConv &           1 x P x T &            4 x P x T &             4 x 1 x 1 x 25 &       convolution &    padding: same, reflect \\
    SpatConv &           4 x P x T &       40 x 1 x T &           40 x 4 x P x 1 &       convolution &            padding: valid \\
    CovPool &      40 x T &    40 x 40 &                    &  covariance  &  temporal dimension \\
    BiMap &   40 x 40 &    20 x 20 &  40 x 20           &            bilinear &       subspace projection \\
    ReEig &   20 x 20 &    20 x 20 &                    &        EV threshold &        threshold = 0.0001 \\
    SPDDSMBN &   20 x 20&    20 x 20 &                  1\tablefootnote{Shared Frechet variance parameter $\nu_{\phi}^2$.} &         TSM         &  domain alignment                         \\
    LogEig &   20 x 20 &              210 &                   &  TSM                &                           \\
    Linear &             210 &        C &    211 x C &              linear &        softmax activation \\
    \bottomrule
    \end{tabular}

\end{table}

\subsection{Supplementary results}

\subsubsection{EEG data}
\label{appendix:results:eeg}

The test set score for each considered EEG dataset is summarized in in Table\,\ref{tab:reults:details}.
Significant differences between the proposed method (TSMNet with SPDDSMBN) and baseline methods are highlighted.


\begin{table}
  \caption{Average (standard deviation across sessions or subjects) test set score (balanced accuracy; higher is better) for all BCI datasets and evaluations.
  Permutation-paired t-test were used to identify significant differences between the proposed (\emph{highlighted}) and baseline methods (1e4 permutations, 10 tests, tmax correction).
  Significant differences are highlighted (\sig$p \le 0.05$, \ssig$p \le 0.01$, \sssig$p \le 0.001$).}
  \label{tab:reults:details}
  \footnotesize
  \centering
  \begin{tabular}{llrrrr}
    \toprule
        & dataset &  \multicolumn{2}{c}{\textbf{BNCI2014001}} & \multicolumn{2}{c}{\textbf{BNCI2015001}} \\
        & evaluation &      inter-session &     inter-subject &      inter-session &      inter-subject \\
        & degrees of freedom / \# classes &                    17 / 4 &                    8 / 4 &                    27 / 2  &                    11 / 2 \\
    UDA & method &                    &                   &                    &                    \\
    \midrule
    no & FBCSP+SVM &    \sig60.6\phantom{0}(\phantom{0}4.9) &  \sig32.3\phantom{0}(\phantom{0}7.3) &   \ssig81.5\phantom{0}(\phantom{0}4.4) &   \ssig58.6\phantom{0}(13.4) \\
    & TSM+SVM &   \ssig61.8\phantom{0}(\phantom{0}4.1) &  \sig34.7\phantom{0}(\phantom{0}8.6) &  \sssig75.7\phantom{0}(\phantom{0}5.1) &  \sssig56.0\phantom{0}(\phantom{0}6.0) \\
    & FB+TSM+LR &       \phantom{0}69.8\phantom{0}(\phantom{0}4.8) &  \sig36.5\phantom{0}(\phantom{0}8.2) &    \sig80.9\phantom{0}(\phantom{0}6.0) &   \ssig60.6\phantom{0}(10.9) \\
    & EEGNet &  \sssig41.8\phantom{0}(\phantom{0}5.8) &  \sig43.3\phantom{0}(17.0) &  \sssig72.4\phantom{0}(\phantom{0}8.4) &  \sssig59.2\phantom{0}(\phantom{0}9.5) \\
    & ShConvNet &  \sssig51.3\phantom{0}(\phantom{0}2.3) &  \sig42.2\phantom{0}(16.2) &  \sssig74.1\phantom{0}(\phantom{0}4.2) &  \sssig58.7\phantom{0}(\phantom{0}5.8) \\
yes & FBCSP+DSS+LDA &       \phantom{0}\textbf{71.3}\phantom{0}(\phantom{0}1.8) &     \phantom{0}48.3\phantom{0}(14.3) &       \phantom{0}84.6\phantom{0}(\phantom{0}4.8) &   \ssig67.7\phantom{0}(14.3) \\
    & URPA+MDM &  \sssig59.5\phantom{0}(\phantom{0}2.7) &     \phantom{0}46.8\phantom{0}(14.6) &  \sssig79.2\phantom{0}(\phantom{0}4.6) &   \ssig70.3\phantom{0}(16.1) \\
    & SPDOT+TSM+SVM &       \phantom{0}66.8\phantom{0}(\phantom{0}3.8) &  \sig38.6\phantom{0}(\phantom{0}8.6) &  \sssig77.5\phantom{0}(\phantom{0}2.9) &  \sssig63.3\phantom{0}(\phantom{0}8.1) \\
    & EEGNet+DANN &  \sssig50.0\phantom{0}(\phantom{0}7.7) &     \phantom{0}45.8\phantom{0}(18.0) &  \sssig71.6\phantom{0}(\phantom{0}5.3) &  \sssig63.7\phantom{0}(11.1) \\
    & ShConvNet+DANN &  \sssig51.6\phantom{0}(\phantom{0}3.2) &  \sig42.2\phantom{0}(13.6) &  \sssig74.1\phantom{0}(\phantom{0}4.0) &  \sssig64.2\phantom{0}(11.6) \\
    & \emph{TSMNet(SPDDSMBN)} &       \phantom{0}69.0\phantom{0}(\phantom{0}3.6) &     \phantom{0}\textbf{51.6}\phantom{0}(16.5) &       \phantom{0}\textbf{85.8}\phantom{0}(\phantom{0}4.3) &       \phantom{0}\textbf{77.0}\phantom{0}(13.7) \\
    \toprule
        & dataset & \multicolumn{2}{c}{\textbf{Lee2019}} & \multicolumn{2}{c}{\textbf{Stieger2021}} \\
        & evaluation &      inter-session &      inter-subject &      inter-session &      inter-subject \\
        & degrees of freedom / \# classes &                    107 / 2 &                    53 / 2 &                    411 / 4 &                    61 / 4\\
    UDA & method &                    &                    &                    &                    \\
    \midrule
    no & FBCSP+SVM &  \sssig63.1\phantom{0}(\phantom{0}4.2) &  \sssig63.4\phantom{0}(12.1) &  \sssig47.5\phantom{0}(\phantom{0}7.0) &  \sssig37.6\phantom{0}(10.5) \\
    & TSM+SVM &  \sssig62.5\phantom{0}(\phantom{0}3.3) &  \sssig65.3\phantom{0}(13.0) &  \sssig49.5\phantom{0}(\phantom{0}8.1) &  \sssig40.2\phantom{0}(12.3) \\
    & FB+TSM+LR &   \ssig65.2\phantom{0}(\phantom{0}4.5) &  \sssig68.5\phantom{0}(12.4) &  \sssig57.3\phantom{0}(\phantom{0}7.3) &  \sssig40.3\phantom{0}(\phantom{0}9.2) \\
    & EEGNet &  \sssig51.2\phantom{0}(\phantom{0}2.7) &  \sssig69.6\phantom{0}(13.8) &  \sssig58.3\phantom{0}(\phantom{0}7.9) &   \ssig43.1\phantom{0}(11.0) \\
    & ShConvNet &  \sssig57.8\phantom{0}(\phantom{0}4.0) &  \sssig68.5\phantom{0}(13.6) &  \sssig60.1\phantom{0}(\phantom{0}6.6) &  \sssig42.2\phantom{0}(10.4) \\
yes & FBCSP+DSS+LDA &       \phantom{0}66.8\phantom{0}(\phantom{0}4.1) &  \sssig68.7\phantom{0}(13.8) &  \sssig59.4\phantom{0}(\phantom{0}6.6) &   \ssig48.2\phantom{0}(13.4) \\
    & URPA+MDM &  \sssig63.8\phantom{0}(\phantom{0}4.2) &  \sssig66.7\phantom{0}(12.3) &  \sssig47.0\phantom{0}(\phantom{0}6.6) &  \sssig38.7\phantom{0}(10.4) \\
    & SPDOT+TSM+SVM &   \ssig65.6\phantom{0}(\phantom{0}4.2) &  \sssig65.4\phantom{0}(10.5) &  \sssig50.3\phantom{0}(\phantom{0}5.8) &  \sssig42.1\phantom{0}(10.5) \\
    & EEGNet+DANN &  \sssig55.4\phantom{0}(\phantom{0}4.4) &  \sssig69.4\phantom{0}(13.1) &  \sssig60.1\phantom{0}(\phantom{0}6.9) &   \ssig43.6\phantom{0}(10.7) \\
    & ShConvNet+DANN &  \sssig59.1\phantom{0}(\phantom{0}3.4) &  \sssig66.0\phantom{0}(12.4) &  \sssig61.3\phantom{0}(\phantom{0}6.0) &  \sssig43.1\phantom{0}(11.5) \\
    & \emph{TSMNet(SPDDSMBN)} &       \phantom{0}\textbf{68.2}\phantom{0}(\phantom{0}4.1) &       \phantom{0}\textbf{74.6}\phantom{0}(14.2) &       \phantom{0}\textbf{64.8}\phantom{0}(\phantom{0}6.8) &       \phantom{0}\textbf{48.9}\phantom{0}(14.3) \\
    \toprule
        & dataset &     \textbf{Lehner2021} &    \textbf{Hehen.2021} & \multicolumn{2}{c}{\textbf{Hinss2021}} \\
        & evaluation &    inter-session &      inter-session &      inter-session &      inter-subject \\
        & degrees of freedom / \# classes &                    6 / 2 &                    25 / 4 &                    29 / 3 &                    14 / 3 \\
    UDA & method &                  &                    &                    &                    \\
    \midrule
    no & FBCSP+SVM &     \phantom{0}68.9\phantom{0}(\phantom{0}6.0) &   \ssig52.5\phantom{0}(\phantom{0}7.1) &   \ssig43.7\phantom{0}(\phantom{0}8.2) &       \phantom{0}45.6\phantom{0}(\phantom{0}6.5) \\
    & TSM+SVM &     \phantom{0}62.7\phantom{0}(\phantom{0}9.1) &  \sssig44.8\phantom{0}(\phantom{0}7.3) &  \sssig36.8\phantom{0}(\phantom{0}4.5) &   \ssig41.7\phantom{0}(\phantom{0}7.4) \\
    & FB+TSM+LR &     \phantom{0}73.0\phantom{0}(\phantom{0}9.6) &   \ssig52.2\phantom{0}(\phantom{0}6.0) &  \sssig40.8\phantom{0}(\phantom{0}7.1) &    \sig45.1\phantom{0}(\phantom{0}5.0) \\
    & EEGNet &  \sig49.6\phantom{0}(\phantom{0}6.4) &  \sssig48.2\phantom{0}(\phantom{0}6.3) &   \ssig46.3\phantom{0}(10.1) &       \phantom{0}47.8\phantom{0}(\phantom{0}5.1) \\
    & ShConvNet &  \sig56.3\phantom{0}(\phantom{0}7.3) &  \sssig53.0\phantom{0}(\phantom{0}5.1) &       \phantom{0}48.9\phantom{0}(\phantom{0}7.4) &    \sig45.9\phantom{0}(\phantom{0}6.8) \\
yes & FBCSP+DSS+LDA &     \phantom{0}77.1\phantom{0}(\phantom{0}8.4) &       \phantom{0}56.4\phantom{0}(\phantom{0}5.3) &   \ssig47.1\phantom{0}(\phantom{0}7.4) &       \phantom{0}48.4\phantom{0}(\phantom{0}9.0) \\
    & URPA+MDM &     \phantom{0}70.8\phantom{0}(\phantom{0}8.2) &  \sssig46.6\phantom{0}(\phantom{0}7.2) &       \phantom{0}51.4\phantom{0}(\phantom{0}3.7) &       \phantom{0}48.4\phantom{0}(\phantom{0}6.1) \\
    & SPDOT+TSM+SVM &  \sig63.0\phantom{0}(\phantom{0}9.2) &  \sssig45.9\phantom{0}(\phantom{0}6.0) &  \sssig42.0\phantom{0}(\phantom{0}4.7) &  \sssig40.4\phantom{0}(\phantom{0}7.5) \\
    & EEGNet+DANN &  \sig49.8\phantom{0}(\phantom{0}3.7) &  \sssig49.3\phantom{0}(\phantom{0}6.7) &    \sig49.4\phantom{0}(\phantom{0}6.8) &       \phantom{0}50.0\phantom{0}(\phantom{0}7.3) \\
    & ShConvNet+DANN &  \sig57.5\phantom{0}(\phantom{0}7.6) &    \sig54.0\phantom{0}(\phantom{0}5.1) &       \phantom{0}51.5\phantom{0}(\phantom{0}4.9) &       \phantom{0}48.8\phantom{0}(\phantom{0}5.7) \\
    & \emph{TSMNet(SPDDSMBN)} &     \phantom{0}\textbf{77.7}\phantom{0}(10.0) &       \phantom{0}\textbf{57.8}\phantom{0}(\phantom{0}5.8) &       \phantom{0}\textbf{54.7}\phantom{0}(\phantom{0}7.3) &       \phantom{0}\textbf{52.4}\phantom{0}(\phantom{0}8.8) \\
    \bottomrule
    \end{tabular}
    
\end{table}

\end{document}